\documentclass{article} 
\usepackage{2026_conference,times}

\usepackage{amsmath,amsfonts,bm}

\def\eqref#1{equation~\ref{#1}}

\def\1{\bm{1}}

\DeclareMathAlphabet{\mathsfit}{\encodingdefault}{\sfdefault}{m}{sl}
\SetMathAlphabet{\mathsfit}{bold}{\encodingdefault}{\sfdefault}{bx}{n}

\usepackage{hyperref}
\usepackage{url}

\usepackage{algorithm}
\usepackage{algorithmic}
\usepackage{xcolor}
\usepackage{colortbl}
\usepackage{amsmath}
\usepackage{amssymb}
\usepackage{booktabs}
\usepackage{graphicx}
\usepackage{wrapfig}
\usepackage{subcaption}
\usepackage{enumitem}

\title{BideDPO: Conditional Image Generation with Simultaneous Text and Condition Alignment}

\author{
    \textbf{Dewei Zhou$^{1}$, Mingwei Li$^{1}$, Zongxin Yang$^{2}$, Yu Lu$^{1}$,} \\
    \textbf{Yunqiu Xu$^{1}$, Zhizhong Wang$^{3}$, Zeyi Huang$^{3}$, Yi Yang$^{1, \ast}$} \\[2mm]
    $^{1}$Zhejiang University \quad $^{2}$Harvard University \quad $^{3}$Huawei Technologies Ltd., China
}

\newcommand{\rebuttal}[1]{{#1}}

\iclrfinalcopy 
\begin{document}

\maketitle
\renewcommand{\thefootnote}{\fnsymbol{footnote}} 
\footnotetext[1]{Corresponding author.} 
\renewcommand{\thefootnote}{\arabic{footnote}} 

\begin{figure}[h]
	\includegraphics[width=1.0\linewidth]{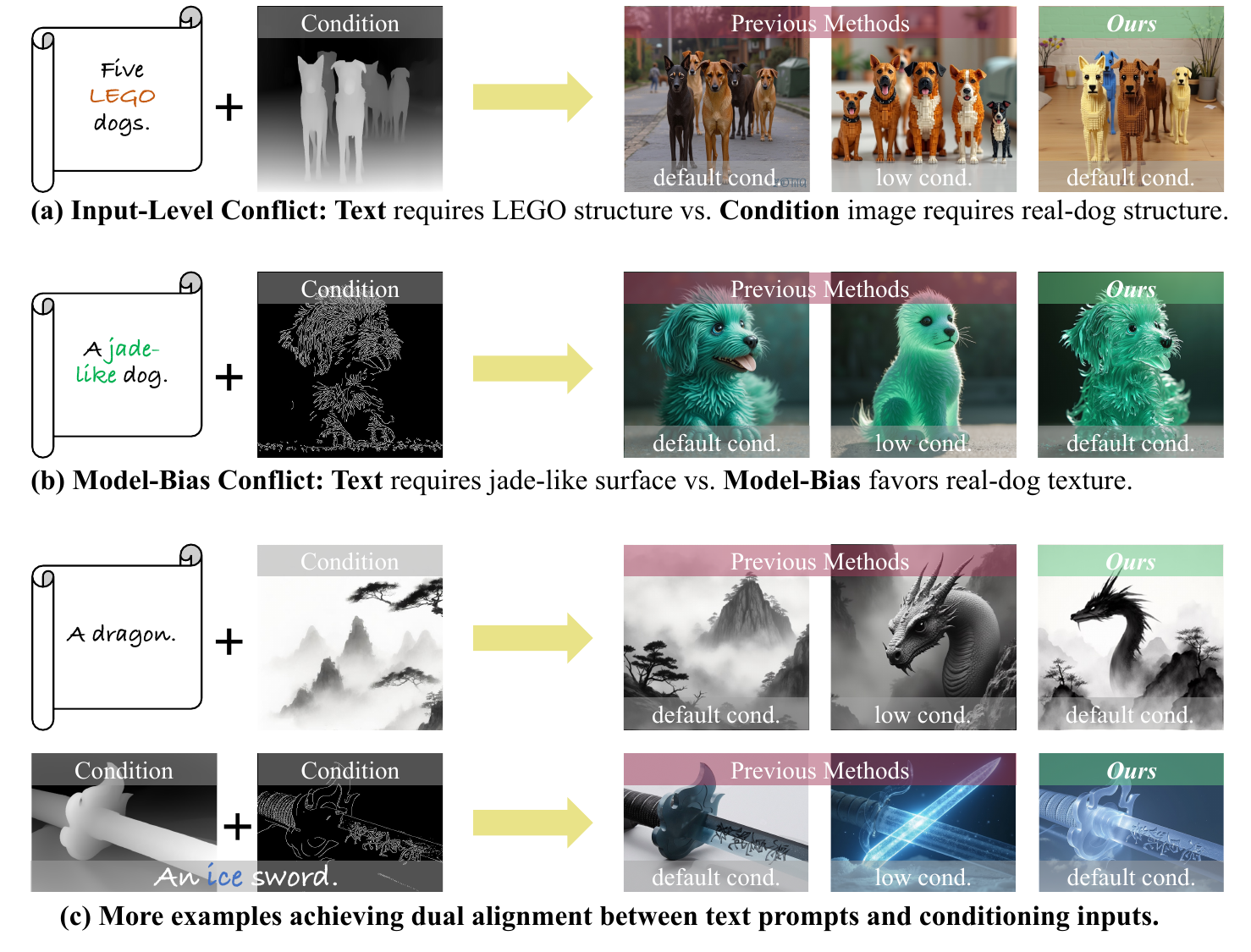}
	\vspace{-6mm}
	\caption{
		\textbf{Qualitative comparison on cases with conflicting text and condition.} 
        We first introduce two conflicts between the text prompt and conditioning input: (a) Input Level Conflict and (b) Model Bias Conflict, which hinder model controllability. We then propose a solution that resolves both, generating images that satisfy both the text and the condition. ``default cond.'' means using its default condition constraint scale, while ``low cond.'' means using a lower condition constraint scale.
        \rebuttal{(c) Our method also enhances the alignment between text and abstract conditions such as style condition, and supports generation with multiple conditions combined with text prompts.}
	}
	\label{fig:effect_vis}

\end{figure}

\begin{abstract}
\rebuttal{Conditional image generation augments text-to-image synthesis with structural, spatial, or stylistic priors and is used in many domains.}
However, current methods struggle to harmonize guidance from both sources when conflicts arise: 1) input-level conflict, where the semantics of the conditioning image contradict the text prompt, and 2) model-bias conflict, where learned generative biases hinder alignment even when the condition and text are compatible. 
These scenarios demand nuanced, case-by-case trade-offs that standard supervised fine-tuning struggles to deliver.
Preference-based optimization techniques, such as Direct Preference Optimization (DPO), offer a promising solution but remain limited: naive DPO suffers from gradient entanglement between text and condition signals and lacks disentangled, conflict-aware training data for multi-constraint tasks.
To overcome these issues, we propose a self-driven, bidirectionally decoupled DPO framework (BideDPO). 
At its core, our method constructs two disentangled preference pairs for each sample—one for the condition and one for the text—to mitigate gradient entanglement. 
The influence of these pairs is then managed by an Adaptive Loss Balancing strategy for balanced optimization. 
To generate these pairs, we introduce an automated data pipeline that iteratively samples from the model and uses vision-language model checks to create disentangled, conflict-aware data. Finally, this entire process is embedded within an iterative optimization strategy that progressively refines both the model and the data.
We construct a DualAlign benchmark to evaluate a model’s ability to resolve conflicts between text and condition, and experiments on commonly used modalities show that BideDPO delivers substantial gains in both text success rate (e.g., +35\%) and condition adherence. We also validated the robustness of our approach on the widely used COCO dataset.
Project Pages: https://limuloo.github.io/BideDPO/.
\end{abstract}
\section{Introduction}
Conditional image generation~\citep{controlnet,li2024controlnet++,liu2024smartcontrol,zavadski2024controlnetxs} augments text-to-image synthesis with auxiliary constraints (\textit{e.g.}, structural or spatial priors) and is now widely used in digital art, design, and related workflows. 
However, real-world use with complex prompt–condition pairs reveals a fundamental yet underexplored challenge: reconciling guidance from the text prompt with the conditioning input. \textbf{In this work, we are the first to explicitly identify this problem and propose a solution.} Specifically, we highlight two recurrent conflicts that undermine model controllability:
\textbf{1) Input-Level Conflict.} When the condition image contains strong semantics that contradict the user prompt, current models often fail to balance these competing sources of guidance. As shown in Fig.~\ref{fig:effect_vis}(a), under the official default setting, models typically prioritize the condition image, resulting in outputs that closely replicate its semantics while neglecting the prompt. Conversely, weakening the influence of the condition allows the model to better follow the text, but often at the expense of spatial or structural consistency. \textbf{2) Model-Bias Conflict.} Modern conditional generation models possess strong generative bias—that is, given a particular condition input, the model tends to produce outputs consistent with its learned biases. As illustrated in Fig.~\ref{fig:effect_vis}(b), even when the condition and text are theoretically compatible, a mismatch between the model's prior and the prompt can lead to poor adherence to the textual guidance.
Addressing the above conflicts requires the model to effectively navigate trade-offs between the condition input and the text prompt, for which no universally optimal solution exists.
In this context, rather than directly providing the model with fixed ``correct" outputs through supervised learning, a more flexible and effective alternative is to guide the model using preference data—examples that reflect human judgments over competing outputs.
Motivated by this, we adopt the Direct Preference Optimization (DPO)~\citep{directpreferenceoptimization} approach, which has been shown to effectively align model outputs with human preferences in both large language models and standard text-to-image generation, and apply it to train our model to better resolve conflicts between condition inputs and text prompts based on preference data.

\begin{figure*}[tb!]
	\includegraphics[width=1.0\linewidth]{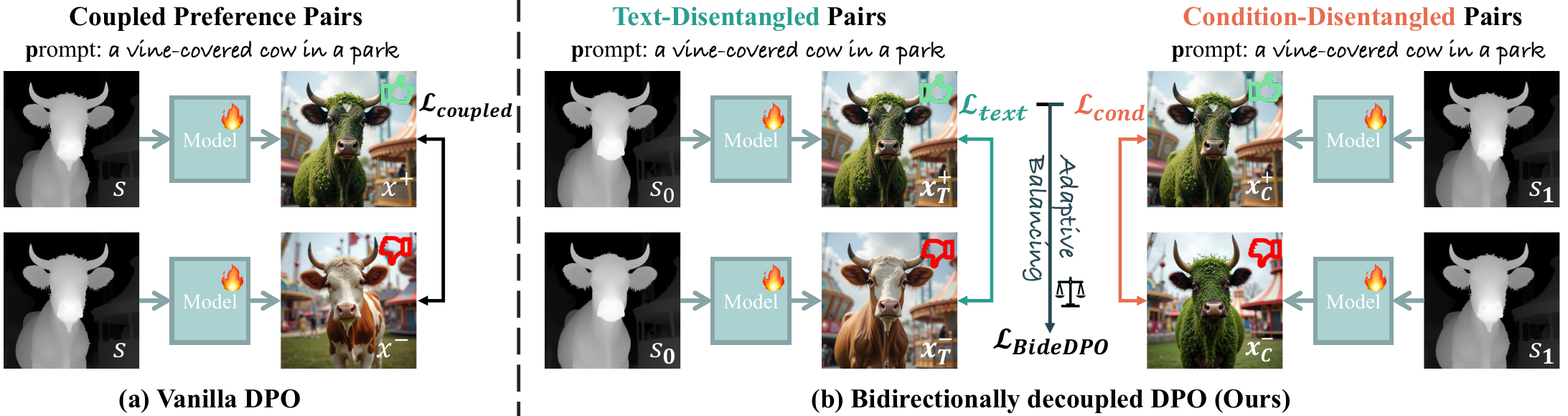}
	\vspace{-6mm}
	\caption{
		\textbf{Comparison between vanilla DPO and our bidirectionally decoupled DPO for conditional image generation.} a) Vanilla DPO uses coupled preference pairs, so its gradients can become ambiguous or even vanish when text and condition are not aligned together.
b) BideDPO separates the learning signals for text and condition and adaptively balance them. This provides clear, adaptive gradients for each requirement, allowing the model to achieve better multi-constraint alignment.
	}
	\vspace{-6mm}
	\label{fig:vs_DPO}
\end{figure*}

However, introducing DPO into the image-conditioned generation task proves to be highly non-trivial, presenting two major challenges. \textbf{1) Naive DPO fails to achieve balanced alignment of both constraints.} In naive DPO, a single preference pair is used for each example. To jointly improve both condition and text alignment for each case, it is necessary to set the positive sample to satisfy both constraints and the negative sample to satisfy neither (Fig.~\ref{fig:vs_DPO}(a)). However, we observe that the model often prioritizes the condition input while neglecting the text, especially when these guidance signals conflict  (see Fig.~\ref{fig:result_vis}). This limitation stems from gradient entanglement, as the coupled learning signals obscure the optimization direction for the weaker constrain, making it difficult for the model to balance and improve both constraints simultaneously. \textbf{2) Lack of Disentangled, Conflict-Aware Preference DPO Data:} To the best of our knowledge, there are no established DPO datasets tailored for conditional image generation, especially for scenarios where the condition input and text prompt provide conflicting or competing guidance. This data gap significantly limits the exploration and benchmarking of preference-based optimization in multi-constraint settings.

To overcome these challenges, we propose a \textbf{self-driven, bidirectionally decoupled DPO framework} tailored for image-conditioned generation. Our framework consists of three key components: \textbf{1) Bidirectionally decoupled DPO algorithm (BideDPO)}: As shown in Fig~\ref{fig:vs_DPO}(b), unlike naive DPO, our method constructs two decoupled preference pairs per example—one for condition fidelity and one for text adherence—used simultaneously during optimization. 

An adaptive loss balancing strategy further ensures balanced progress on both objectives, preventing the model from collapsing toward a single constraint. 
This decoupling, combined with dynamic loss adjustment, enables the model to better handle conflicts between the two constraints and achieve a more effective trade-off.
\textbf{2) Automated Construction of Disentangled and Conflict-Aware DPO Preference Data}: As shown in Fig.~\ref{fig:data_pipe}, we address the lack of suitable DPO data by introducing an automated pipeline that iteratively samples from the current model and uses vision-language model (VLM) checks to construct high-quality positive and negative samples for both text and condition branches. These datasets include numerous instances where the text and the condition are in conflict.
\textbf{3) An iterative optimization strategy}: 
Our framework naturally supports iterative refinement because the generator itself produces the preference data used for training.
As shown in Fig.~\ref{fig:overview}, we alternate between generating preference pairs with the current model and optimizing it with BideDPO. Each round leverages the improved generator to produce higher-quality data, creating a self-reinforcing loop that progressively enhances both model performance and data quality. 

We construct a DualAlign benchmark to evaluate how state-of-the-art conditional image generation methods handle conflicts between the conditioning input and the text prompt. Experiments are conducted across standard conditioning modalities. Results show that our method markedly improves the text success rate (SR) and adherence to the conditioning signal over strong baselines—for example, on FLUX-Depth it boosts text SR by \textbf{15}\% and reduces the conditional MSE by \textbf{87.7} points. We also validate robustness on the standard COCO benchmark~\citep{coco,zhang2025easycontrol}: even with standard prompts, our approach delivers substantial gains over the original model, e.g., a \textbf{15}\% improvement for canny-conditioned generation. Furthermore, our iterative optimization strategy proves effective: as the number of iterations increases, the model consistently achieves better trade-offs between conditioning and text, yielding steady performance gains.

We summarize our contributions as follows:
\begin{itemize}[leftmargin=*]
    \item To the best of our knowledge, this work is the first to formally formulate and systematically analyze the text–condition adherence conflict in conditional image generation. We propose BideDPO, an effective DPO algorithm that reconciles conflicts between condition inputs and text prompts.

    \item We propose an automated pipeline that produces disentangled condition–text preference pairs. It easily extends to other tasks and supports iterative optimization, improving both model and data.

    \item We construct a DualAlign Benchmark for evaluating a model’s ability to handle conflicts between condition inputs and text prompts, and the results highlight the effectiveness of our approach.
\end{itemize}

\section{Related Work}
\noindent\textbf{Conditional Image Generation.}
With the rapid advancement of image generation technology~\citep{stablediffusion, podell2023sdxl, SD2, esser2024sd3,lu2023tf,lu2024mace,lu2025does,lu2024robust,zhou2025dragflow,pydiff,migc,migc++,zhou20243dis,zhou20253disflux,zhou2025dreamrenderer,zhao2023wavelet,zhao2024learning,zhao2025zero,zhao2025ultrahr,zhao2024toward}, current models are now capable of producing highly realistic images. As a result, a key research focus has shifted toward enabling more conditional image generation~\citep{li2024controlnet++, bhat2024loosecontrol,lin2024ctrladapter,peng2024controlnext, wang2024omnicontrolnet,ip-adapter,zhang2025easycontrol}, 
aiming to make text-to-image technologies applicable in a broader range of real-world scenarios. 
For example, models of the ControlNet~\citep{controlnet,xu2024ctrlora,zhao2023unicontrol} family, such as FLUX.1-dev-UnionPro2~\citep{unionpro2}, introduce an auxiliary network to inject user-provided spatial constraints or reference image information into the generation process. Other variants, including FLUX-Depth and FLUX-Canny~\citep{flux}, concatenate the encoded condition image with the generated image features at the input channel, training the entire network end-to-end to achieve conditional image generation.

\noindent\textbf{Aligning Image Generation with Human Preferences.}
Reinforcement learning (RL) methods are widely used for post-training large language models (LLMs) to align outputs with human preferences, and has recently been applied to image generation for improved controllability and preference alignment. Most approaches rely on explicit reward models, such as ImageReward~\citep{xu2023imagereward}, combined with policy rollouts like PPO~\citep{liu2025flowgrpo,dancegrpo} or direct gradient methods~\citep{clark2024draft,prabhudesai2023alignprop}. A more efficient alternative is Direct Preference Optimization (DPO)~\citep{directpreferenceoptimization}, adapted to diffusion models by Diffusion-DPO~\citep{diffusiondpo}, and further extended for richer feedback (RankDPO~\citep{rankdpo}) and timestep inconsistencies (SPO~\citep{liang2024spo}, TailorPO~\citep{tailorpo}). Given its efficiency and effectiveness, we build on the DPO framework. However, existing DPO-based methods still struggle with conditional image generation involving multiple, potentially conflicting constraints. To address this, we propose a novel \textbf{Bidirectionally Decoupled DPO} algorithm that enables clearer optimization directions and better handles complex conditional scenarios.
\section{Method}
Our method has three components: (1) a Bidirectionally Decoupled DPO algorithm (BideDPO, \S\ref{sec:bidedpo}) that enforces simultaneous text–condition adherence; (2) an automatic pipeline for constructing Disentangled and Conflict-Aware Preference Data (\S\ref{sec:data_pipe}) for BideDPO training; and (3) an iterative optimization strategy (\S\ref{sec:iter_opt}) that jointly improves model performance and data quality.
\textbf{Preliminary background on Diffusion Models and DPO is provided in Appendix~\S\ref{sec:preliminary}.}

\subsection{Bidirectionally Decoupled DPO}
\label{sec:bidedpo}
\noindent\textbf{Limitations of Vanilla DPO.}
In conditional image generation, models are often tasked with satisfying both a text prompt $p$ and an extra structural condition $s$. For notational simplicity, we encapsulate both inputs into a single composite context variable $c = (p, s)$. To analyze the optimization dynamics of DPO~\citep{directpreferenceoptimization} in this multi-objective setting, we can conceptualize the model's preference score as being composed of a text alignment component, $f_{\text{text}}(x, c; \theta)$ (which primarily depends on $p$), and a condition alignment component, $f_{\text{cond}}(x, c; \theta)$ (which primarily depends on $s$).

For simplicity, let us assume the overall score $f(x, c; \theta)$ can be modeled as a weighted linear combination of these components, while acknowledging that the true relationship is far more complex:
\begin{equation}
f(x, c; \theta) =  \lambda_{\text{text}} f_{\text{text}}(x, c; \theta) +  \lambda_{\text{cond}} f_{\text{cond}}(x, c; \theta),
\label{eq:score}
\end{equation}
where $\lambda_{\text{text}}$ and $\lambda_{\text{cond}}$ are scalar weights. For a preference triplet $(x^+, x^-, c)$, where both the preferred and dispreferred samples are evaluated under the same shared composite context $c$, the vanilla DPO loss is:
{
\begin{equation}
\mathcal{L}_{\text{coupled}} = -\log \sigma(f(x^+, c; \theta) - f(x^-, c; \theta)).
\label{eq:coupled_loss}
\end{equation}
}

The partial derivative of this loss with respect to the model parameters $\theta$ is given by:
{
\begin{equation}
\frac{\partial \mathcal{L}_{\text{coupled}}}{\partial \theta} = -\left(1 - \sigma(\Delta(c; {\theta})\right) \frac{\partial \Delta(c;{\theta})}{\partial \theta}, \label{eq:coupled_grad}
\end{equation}
}
{
\begin{equation}
\Delta(c;{\theta}) = \lambda_{\text{text}} \underbrace{\left(f_{\text{text}}(x^+, c; \theta) - f_{\text{text}}(x^-, c; \theta)\right)}_{\Delta_{\text{text}}(c; \theta)} 
 + \lambda_{\text{cond}} \underbrace{\left(f_{\text{cond}}(x^+, c; \theta) - f_{\text{cond}}(x^-, c; \theta)\right)}_{\Delta_{\text{cond}}(c; \theta)}.
\label{eq:delta_combined}
\end{equation}
}
The update gradient in Eq.~\ref{eq:coupled_grad} is influenced simultaneously by both objectives, but when one objective’s gradient is significantly stronger, it can dominate the update. Consequently, the weaker objective may be masked—or, in cases of conflict, the update may even move in a direction that opposes the weaker objective, making its optimization more difficult.

\noindent\textbf{Decoupled Preference Pairs.}
To address this limitation, we construct two decoupled preference pairs for each case, targeting condition and text alignment separately (Fig.~\ref{fig:vs_DPO}(b)). For text alignment, we use a pair $(x_T^+, x_T^-, c_0)$, and for condition alignment, we use $(x_C^+, x_C^-, c_1)$. Here, $c_0 = (p, s_0)$ and $c_1 = (p, s_1)$, where $p$ is the same target prompt in both cases. In the text alignment pair, $x_T^+$ and $x_T^-$ have similar adherence to $s_0$, but $x_T^+$ follows the prompt $p$, while $x_T^-$ does not. In the condition alignment pair, both $x_C^+$ and $x_C^-$ follow $p$, but $x_C^+$ matches $s_1$ much better than $x_C^-$.
Two independent loss terms are then calculated separately for each objective:
{
\begin{align}
\mathcal{L}_{\text{text}} &= -\log \sigma\underbrace{\left(f_{\text{text}}(x^+_T, c_0; \theta) - f_{\text{text}}(x^-_T, c_0; \theta)\right)}_{\Delta_{\text{text}}(c_0; \theta)}, \\
\mathcal{L}_{\text{cond}} &= -\log \sigma\underbrace{\left(f_{\text{cond}}(x^+_C, c_1; \theta) - f_{\text{cond}}(x^-_C, c_1; \theta) \right)}_{\Delta_{\text{cond}}(c_1; \theta)}
\end{align}
}

\noindent\textbf{Adaptive Loss Balancing.} 
To prevent the optimization from being dominated by one objective, we introduce an adaptive loss balancing strategy. To ensure stable training, the weights for each loss component are computed based on their current magnitudes but are treated as detached constants during backpropagation. This is achieved by applying a stop-gradient operator, denoted as $\text{sg}(\cdot)$:
{
\begin{equation}
    w_{\text{text}} = \text{sg}\left(\frac{\mathcal{L}_{\text{text}}}{\mathcal{L}_{\text{text}} + \mathcal{L}_{\text{cond}}}\right) ,
     \text{ and} \quad w_{\text{cond}} = \text{sg}\left(1 - w_{\text{text}}\right).
\end{equation}
}
The total loss is thus a dynamically weighted sum:
{
\begin{equation}
    \mathcal{L}_{\text{decoupled}} = w_{\text{text}}\mathcal{L}_{\text{text}} + w_{\text{cond}}\mathcal{L}_{\text{cond}}.
\end{equation}
}
\noindent\textbf{Decoupled gradient.} The gradient can be formulated as: 
\begin{equation}
\begin{split}
\frac{\partial\mathcal{L}_{\text{decoupled}}}{\partial\theta}
&= w_{\text{text}}\frac{\partial\mathcal{L}_{\text{text}}}{\partial\theta}
   + w_{\text{cond}}\frac{\partial\mathcal{L}_{\text{cond}}}{\partial\theta} \\
&\mathrel{\mkern-80mu}= -\,w_{\text{text}}\left(1-\sigma\bigl(\Delta_{\text{text}}(c_0;\theta)\bigr)\right)
  \frac{\partial\Delta_{\text{text}}(c_0;\theta)}{\partial\theta} -\,w_{\text{cond}}\left(1-\sigma\bigl(\Delta_{\text{cond}}(c_1;\theta)\bigr)\right)
  \frac{\partial\Delta_{\text{cond}}(c_1;\theta)}{\partial\theta}.
\end{split}
\end{equation}
Crucially, this gradient is a fully decoupled sum. Unlike the coupled gradient in Eq.~\ref{eq:coupled_grad}, our approach provides a distinct optimization signal for each objective. This prevents one objective's gradient from being diminished or ``swallowed'' when the other's loss is significantly larger, ensuring both are consistently optimized.
\begin{figure*}[tb!]
	\includegraphics[width=1.0\linewidth]{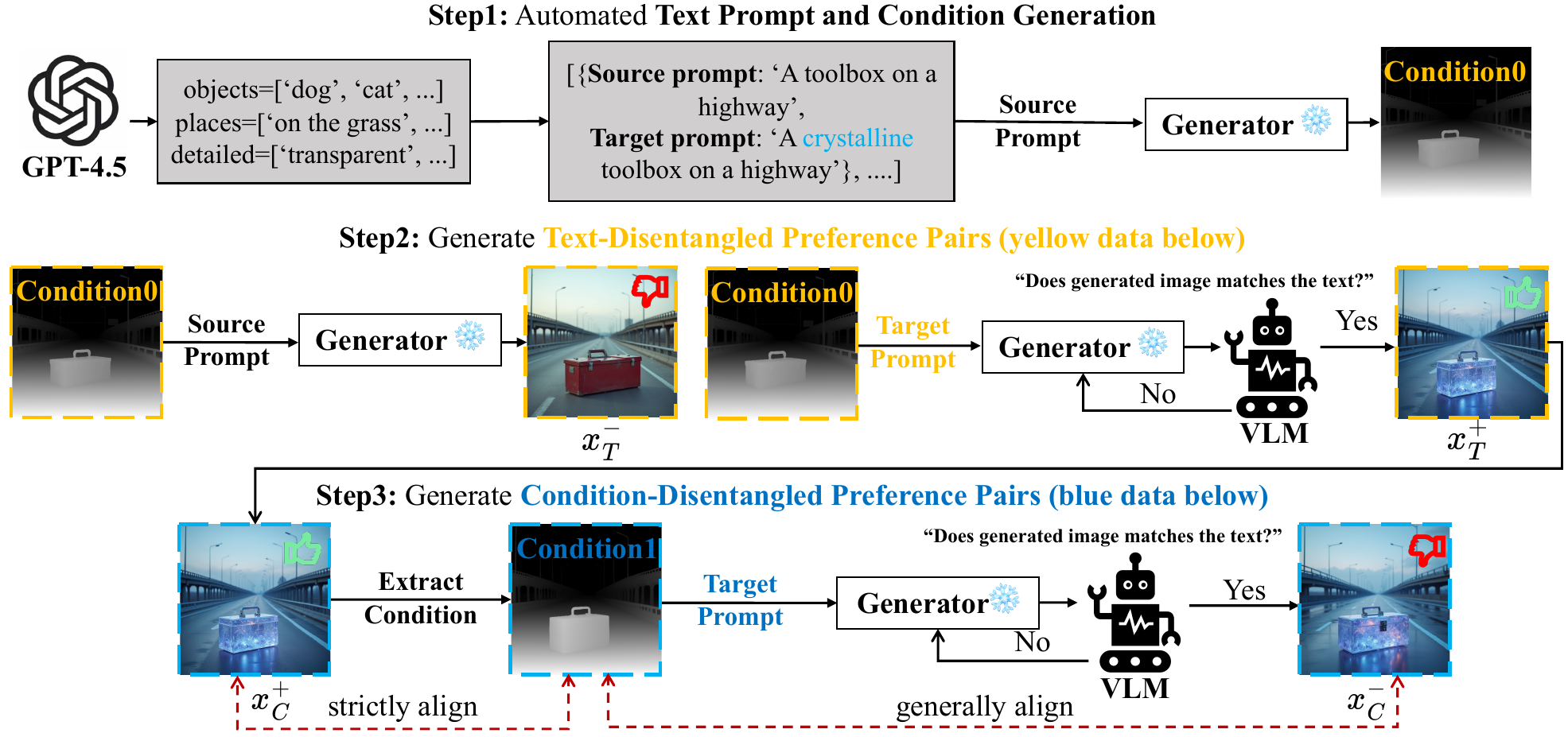}
	\vspace{-6mm}
	\caption{
		\textbf{The Automated Disentangled, Conflict-Aware Preference Data Generation Pipeline.} 
	}
	\vspace{-4mm}
	\label{fig:data_pipe}
\end{figure*}

\noindent\textbf{BideDPO Objective for Diffusion Models.}
Building upon prior work~\citep{diffusiondpo}, we define the reward $r(x_t, c, \epsilon; \theta)$ in diffusion models as the reduction in denoising error for a noisy sample $x_t$ under a given context $c$, which is computed as the difference between the denoising error $\lVert\epsilon - \epsilon_{\text{ref}}(x_t, c)\rVert^2$ of the frozen reference network and the error $\lVert\epsilon - \epsilon_\theta(x_t, c)\rVert^2$ of the optimized network, where $\epsilon$ represents the noise:
{
\begin{equation}
\label{eq:reward_def}
r(x_t, c, \epsilon; \theta) = \lVert\epsilon - \epsilon_{\text{ref}}(x_t, c)\rVert^2 - \lVert\epsilon - \epsilon_\theta(x_t, c)\rVert^2.
\end{equation}
}
The total reward differences for the text pair under context $c_0$ ($R_T$) and the context pair under context $c_1$ ($R_C$) are then:
{
\begin{align}
R_T &= r(x_{t,T}^+, c_0, \epsilon_T^+; \theta) - r(x_{t,T}^-, c_0, \epsilon_T^-; \theta), \\
R_C &= r(x_{t,C}^+, c_1, \epsilon_C^+; \theta) - r(x_{t,C}^-, c_1, \epsilon_C^-; \theta).
\end{align}
}
The final BideDPO loss adaptively weights the objectives based on these reward differences:
{
\begin{equation}
\label{eq:bide-dpo_loss}
\begin{split}
    \mathrel{\mkern-15mu}\mathcal{L}_{\text{BideDPO}}(\theta) = - \mathbb{E}_{\substack{(x_T^+, x_T^-, c_0) \sim \mathcal{D}_T, (x_C^+, x_C^-, c_1) \sim \mathcal{D}_C}} \Big[
        w_{\text{text}} \log\sigma(\beta T  R_T)
        + w_{\text{cond}} \log\sigma(\beta T  R_C)
    \Big].
\end{split}
\end{equation}
}

By providing a distinct gradient for each objective, our decoupled approach mitigates the interference inherent in the decoupled DPO loss. This leads to more stable and efficient multi-objective optimization.
\textbf{Please see Appendix~\S\ref{sec:BideDPO_detail} for more details.}

\subsection{Automated Construction of Disentangled and Conflict-Aware DPO Preference Data}
\label{sec:data_pipe}

As shown in Fig.~\ref{fig:data_pipe}, we design an automated data construction pipeline that explicitly generates disentangled preference pairs for both text and condition alignment, including cases where the two objectives are in conflict.
It consists of three steps:

\noindent\textbf{1. Prompt and Initial Condition Generation.} We first use an LLM to generate a basic \textit{Source Prompt} and a more detailed \textit{Target Prompt} $p$. The source prompt is then used to produce an initial condition map, ``Condition 0'' ($s_{0}$). The generated $s_{0}$ and target prompt $p$ often exhibit input-level or model-bias conflicts in Fig.~\ref{fig:effect_vis}.

\noindent\textbf{2. Text-Disentangled Pair $(x_T^+, x_T^-, p, s_{0})$.} Both samples adhere to ``Condition 0'' ($s_{0}$). The preferred sample $x_T^+$ is generated from the \textit{Target Prompt}, with its textual alignment verified by a VLM, and serves as our high-quality anchor. The dispreferred sample $x_T^-$ is generated from the \textit{Source Prompt} and thus lacks textual alignment with \textit{Target Prompt}.

\noindent\textbf{3. Condition-Disentangled Pair $(x_C^+, x_C^-, p, s_{1})$.} Both samples align with the \textit{Target Prompt} $p$. The anchor $x_T^+$ serves as the preferred sample $x_C^+$, and a new, strictly aligned condition map ``Condition 1'' ($s_{1}$) is extracted from it. The dispreferred sample $x_C^-$ is then generated to adhere less strictly to ``Condition 1'' while matching the target prompt's semantics.

This structured process allows us to systematically generate preference data that isolates and targets distinct aspects of text and condition alignment. By repeating this process over a large set of prompts and conditions, we built a comprehensive dataset that enables targeted, disentangled optimization for multi-constraint image generation. \rebuttal{Moreover, this approach can be easily extended to other tasks, such as style-text alignment, as illustrated in Fig.~\ref{fig:data_style_pipe}.}
\begin{figure}[tb!]
	\includegraphics[width=1.0\linewidth]{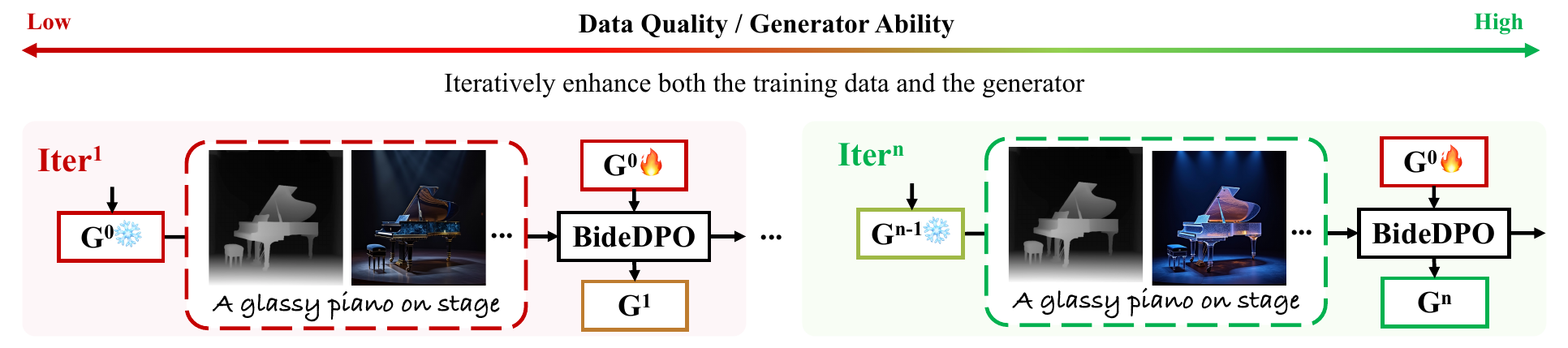}
	\vspace{-6mm}
	\caption{
		\textbf{Iterative Optimization Strategy.} 
        We start with an initial generator ($G^0$) that produces training data via our automated pipeline (Fig.~\ref{fig:data_pipe}). Training with BideDPO already improves the model ($G^1$), while repeating the process with the updated generator yields higher-quality data and further gains, forming a self-reinforcing loop where both data and model improve progressively.
	}
	\vspace{-4mm}
	\label{fig:overview}
\end{figure}

\subsection{Iterative Optimization Strategy}
\label{sec:iter_opt}
For each generator model, BideDPO strengthens adherence to both text and condition, and since our data construction pipeline builds samples directly from the same model, the process naturally supports iterative refinement.
As shown in Fig.~\ref{fig:overview}, we alternate between generating preference data with the current model and optimizing it with BideDPO, forming a self-reinforcing loop that progressively improves both the generator and its training data. 
\section{Experiments}
\subsection{Experimental Setup}

\begin{figure*}[tb!]
	\includegraphics[width=1.0\linewidth]{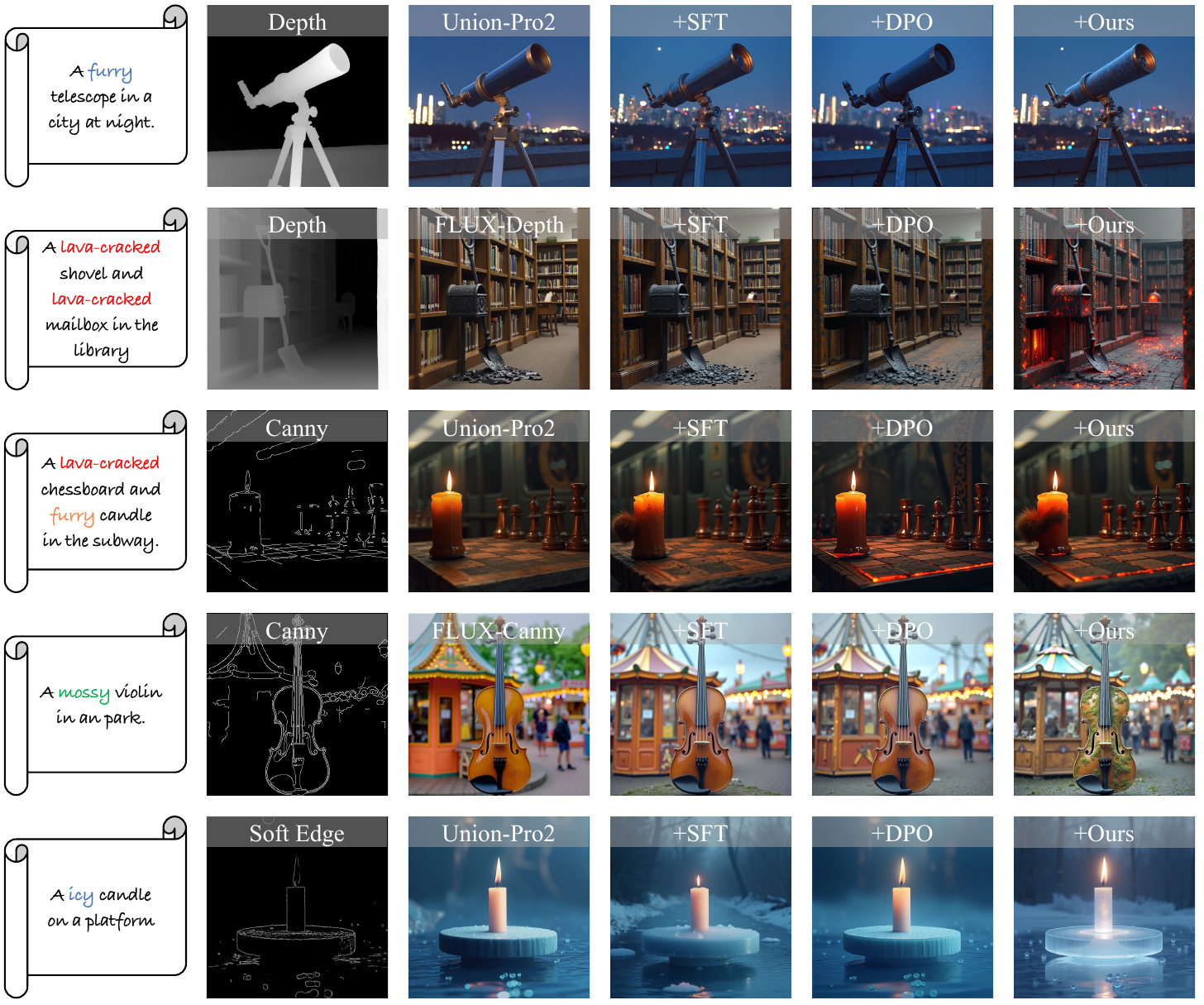}
	\vspace{-6mm}
	\caption{
		\textbf{Visual comparison for conditional image generation on the DualAlign Benchmark.} We evaluate three common conditioning modalities: depth, Canny, and soft edge. Our method improves adherence to both the text prompt and the spatial conditioning. \textbf{Please zoom in for details.}
	}
	\vspace{-4mm}
	\label{fig:result_vis}
\end{figure*}
\noindent\textbf{Baselines.} We conduct experiments on the state-of-the-art text-to-image model FLUX~\citep{flux}. Specifically, we evaluate our approach on the most widely used conditional image generation variants in the community, including FLUX-Depth, FLUX-Canny, and Union-Pro2~\citep{unionpro2}. \rebuttal{We also compare with LooseControl (limited to depth conditioning)~\citep{bhat2024loosecontrol} and ControlNet++~\citep{li2024controlnet++}.} \rebuttal{We evaluate style-conditioned generation on FLUX IP-Apdater~\citep{flux-ipa}.} We primarily compare our method with two common baselines: supervised fine-tuning (SFT) and naive DPO~\citep{diffusiondpo}.

\noindent\textbf{Implementation Details.}
We generate 5,000 samples in each iteration. For SFT, we use the positive samples in Fig.~\ref{fig:data_pipe}. For DPO, we construct coupled preference pairs by combining the condition image and positive sample from the Condition-Disentangled Preference Pairs, together with the negative sample from the Text-Disentangled Preference Pairs. 
We fine-tune all models using Low-Rank Adaptation (LoRA~\citep{hu2022lora}) method with rank of 256. For the SFT method, we train for 5,000 steps using the Prodigy~\citep{mishchenko2023prodigy} optimizer with a learning rate of 1.0, using all positive samples in Fig.~\ref{fig:data_pipe}; for DPO and BideDPO, starting from the SFT-tuned model, we optimize for an additional 2,000 steps using the AdamW~\citep{adam} optimizer with a learning rate of 0.00004 and a weight decay of 0.01.

\noindent\textbf{Evaluation Benchmarks.} \textit{1) DualAlign benchmark for conflicting text–condition constraints.} Currently, there is no established benchmark for evaluating conditional image generation in scenarios where the text prompt and condition image provide partially conflicting guidance. Therefore, we construct our own test set following a similar pipeline as our training data, generating text-condition pairs that require the model to make meaningful trade-offs between constraints. To better assess the generalization ability of our approach, we ensure that the objects, places, and detailed descriptions in the test set do not overlap with those in the training set. Each modality contains 100 cases. \textit{2) COCO benchmark for robustness.} To assess robustness on a standard benchmark, we also evaluate various post-training methods alongside their baseline models on the COCO dataset~\citep{coco,zhang2025easycontrol}, demonstrating that our approach preserves the base model’s original performance. \rebuttal{\textit{3) DualAlign-Style benchmark for text–style condition constraints (see \S ~\ref{sec:style_benchmark} in Appendix).}
}

\begin{table}[t]
  \centering
  \begin{minipage}[t]{0.495\linewidth}
    \centering
    \captionof{table}{\rebuttal{\textbf{Results for depth-conditioned image generation on DualAlign Benchmark.} ``Ctrl.'' indicates support for conditional generation.}}

	\vspace{-4mm}
    
    \label{tab:depth_results}
    \footnotesize

    \centering
        \setlength{\tabcolsep}{0.5mm}
	\begin{tabular}{rccccc}
        \toprule[1pt]
		\textbf{Method}   & \textbf{Ctrl.} & \textbf{SR} $\uparrow$ & \textbf{MSE} $\downarrow$ & \textbf{SGMSE} $\downarrow$ & \textbf{CLIP} $\uparrow$ \\
		\hline
		\rowcolor[HTML]{F5F5F5}
		FLUX                & $\times$ & 0.79                              & N/A                         & N/A                           & 0.2936
                           \\
        		\rowcolor[HTML]{F5F5F5}
		\rebuttal{LooseControl}                & \rebuttal{$\checkmark$} & \rebuttal{0.43}                              & \rebuttal{791.1}                         & \rebuttal{1280.2}                           & \rebuttal{0.2852}
                          \\
        		\rowcolor[HTML]{F5F5F5}
		\rebuttal{ControlNet++}                & \rebuttal{$\checkmark$} & \rebuttal{0.49}                              & \rebuttal{331.9}                         & \rebuttal{480.7}                           & \rebuttal{0.2854}
                          \\
		\hline
		\rowcolor[HTML]{F8FBFF}
		Union-Pro2        & $\checkmark$ & 0.49                              & 177.0
                   & 272.4                     & 0.2748
                           \\
		\rowcolor[HTML]{F8FBFF}
		+ SFT  & $\checkmark$ & 0.70                              & 262.2
                    & 332.5                    & 0.2915
                           \\
		\rowcolor[HTML]{F8FBFF}
		+ DPO  & $\checkmark$ &                0.71                  &  168.3
                         &          219.9                  & 0.2860
                           \\
		\rowcolor[HTML]{F8FBFF}
		+ Ours & $\checkmark$ & \textbf{0.84}                     & \textbf{164.0}           & \textbf{195.7}             & \textbf{0.2924}
                           \\
		\hline
		\rowcolor[HTML]{FFF8F0}
		FLUX-Depth        & $\checkmark$ & 0.76                              & 233.6                   & 282.8                    & 0.2899
                           \\
		\rowcolor[HTML]{FFF8F0}
		+ SFT& $\checkmark$ & 0.79                             & 162.2
                    & 195.7                     & 0.2926
                           \\
		\rowcolor[HTML]{FFF8F0}
		+ DPO  & $\checkmark$ &        0.89                           &           171.9                &                   195.0          & 0.2974
                           \\
		\rowcolor[HTML]{FFF8F0}
		+ Ours & $\checkmark$ & \textbf{0.91}                     & \textbf{145.9}           & \textbf{164.4}             & \textbf{0.2982}
                           \\
		\toprule[1pt]
\end{tabular}
  \end{minipage}\hfill
%
%
\begin{minipage}[t]{0.48\linewidth}
\centering
\captionof{table}{\rebuttal{\textbf{Results for canny-conditioned image generation on DualAlign Benchmark.} ``Ctrl.'' indicates support for conditional generation.}}

\vspace{-4mm}
\footnotesize
    \label{tab:canny_results}
\setlength{\tabcolsep}{1.4mm}%
{

	\centering
        \setlength{\tabcolsep}{0.6mm}
	\begin{tabular}{rccccc}
        \toprule[1pt]
		\textbf{Method}   & \textbf{Ctrl.} & \textbf{SR} $\uparrow$ & \textbf{F1} $\uparrow$ & \textbf{SGF1} $\uparrow$ & \textbf{CLIP} $\uparrow$ \\
		\hline
		\rowcolor[HTML]{F5F5F5}
        FLUX                & $\times$ & 0.71                              & N/A                         & N/A                           & 0.2965   \\
        		\rowcolor[HTML]{F5F5F5}
		\rebuttal{ControlNet++}                & \rebuttal{$\checkmark$} & \rebuttal{0.40}                              & \rebuttal{0.437}                         & \rebuttal{0.174}                           & \rebuttal{0.2828}
                          \\
		\hline
		\rowcolor[HTML]{F8FBFF}
		Union-Pro2        & $\checkmark$ & 0.34
                              & 0.418
                   & 0.143
                     & 0.2753
                           \\
		\rowcolor[HTML]{F8FBFF}
		+ SFT  & $\checkmark$ & 0.58
                              & 0.324
                   & 0.178
                     & 0.2838
                        \\
		\rowcolor[HTML]{F8FBFF}
		+ DPO  & $\checkmark$ & 0.50                                  &           0.607                &                      0.284       & 0.2840
                         \\
		\rowcolor[HTML]{F8FBFF}
		+ Ours & $\checkmark$ & \textbf{0.68}                     & \textbf{0.607}           & \textbf{0.393}             & \textbf{0.2845}                           \\
		\hline
		\rowcolor[HTML]{FFF8F0}
		FLUX-Canny        & $\checkmark$ & 0.33                             & 0.397                    & 0.129                     & 0.2703                           \\
		\rowcolor[HTML]{FFF8F0}
		+ SFT& $\checkmark$ & 0.52                             & 0.357                    & 0.179                    & 0.2842                           \\
		\rowcolor[HTML]{FFF8F0}
		+ DPO  & $\checkmark$ &       0.55                            &                      0.452     &       0.248                      & 0.2829                           \\
		\rowcolor[HTML]{FFF8F0}
		+ Ours & $\checkmark$ & \textbf{0.73}                     & \textbf{0.454}           & \textbf{0.333}             & \textbf{0.2927}                           \\
		\toprule[1pt]
	\end{tabular}
  }  
  \end{minipage}
\end{table}

\noindent\textbf{Evaluation Metrics.} 
We evaluate our models using the following metrics:
\textit{1) Success Ratio:} We use the Qwen2.5-VL-72B~\citep{bai2023qwen} model to automatically determine whether the generated image accurately matches the text description, providing a direct measure of text-image consistency.
\textit{2) CLIP Score~\citep{CLIP}:} This metric quantifies the semantic alignment between the generated image and the input prompt, indicating how well the model captures the intended content described by the user.
\textit{3) MSE/F1 Score:} These metrics assess the degree to which the generated image conforms to the input condition (\textit{e.g.}, spatial or structural constraints), thereby measuring conditional fidelity.
\textit{4) Semantic-Guided MSE (SGMSE), Semantic-Guided F1 (SGF1), and Semantic-Guided SSIM (SGSSIM):} To jointly evaluate textual and conditional alignment, we define SGMSE, SGF1, and SGSSIM. Each extends its standard counterpart by adding a semantic check. If a generation fails the text requirement, we apply a penalty: MSE is doubled, and the F1 or SSIM score is set to zero (otherwise the metrics reduce to the usual MSE, F1, and SSIM). This design penalizes outputs that do not satisfy both constraints and provides a more comprehensive assessment of controllable image generation. 

\subsection{Experimental Results}
\noindent\textbf{Qualitative Results.}
{Fig.~\ref{fig:result_vis} presents visual comparisons between our BideDPO and other post-training methods. Supervised fine-tuning (SFT) reduces the model’s adherence to the input condition. For example, in the fifth row of Fig.~\ref{fig:result_vis}, the shape at the top of the candle is noticeably altered.}
Naive DPO, due to coupled gradients, biases optimization toward condition adherence while often neglecting textual alignment, resulting in outputs that frequently fail to match the text description. 
In contrast, our bidirectionally decoupled and adaptively balanced approach enables the model to resolve conflicts between condition and text better, achieving a more effective trade-off and satisfying both constraints.
Notably, BideDPO and vanilla SFT are trained with exactly the same set of positive examples; this controlled comparison underscores the superiority of our method.

\noindent\textbf{Quantitative Results.}
Quantitative results in Tabs.~\ref{tab:depth_results}, \ref{tab:canny_results}, and \ref{tab:soft_edge_results} show that our method substantially improves adherence to both text prompts and conditioning inputs.
\begin{wraptable}[11]{r}{0.55\linewidth}
\centering
\caption{\textbf{Results for soft edge-conditioned image generation on DualAlign Benchmark.} ``Ctrl.'' indicates support for conditional generation.}
\vspace{-4mm}
\footnotesize
{\setlength{\tabcolsep}{1.mm}%
\begin{tabular}{@{}rccccc@{}}
\toprule
\textbf{Method} & \textbf{Ctrl.} & \textbf{SR} $\uparrow$ & \textbf{SSIM} $\uparrow$ & \textbf{SGSSIM} $\uparrow$ & \textbf{CLIP} $\uparrow$ \\
\midrule
\rowcolor[HTML]{F5F5F5}FLUX        & $\times$     & 0.73  & N/A     & N/A     & 0.2907 \\
\midrule
\rowcolor[HTML]{F8FBFF}Union-Pro2  & $\checkmark$ & 0.24  & 0.610 & 0.145 & 0.2768 \\
\rowcolor[HTML]{F8FBFF}+ SFT       & $\checkmark$ & 0.48  & 0.510 & 0.255 & 0.2855 \\
\rowcolor[HTML]{F8FBFF}+ DPO       & $\checkmark$ & 0.39  & 0.637 & 0.250 & 0.2783 \\
\rowcolor[HTML]{F8FBFF}+ Ours      & $\checkmark$ & \textbf{0.49} & \textbf{0.643} & \textbf{0.297} & \textbf{0.2855} \\
\bottomrule
\end{tabular}%
}
\vspace{-5mm}
\label{tab:soft_edge_results}
\end{wraptable}

For example, in depth-conditioned generation, we observe a \textbf{43}\% increase in Success Ratio on Union-Pro2 and a \textbf{15}\% increase on FLUX-Depth. In canny-conditioned generation, our approach achieves a \textbf{34}\% higher Success Ratio on Union-Pro2 and a \textbf{40}\% increase on FLUX-Canny—even surpassing the original T2I FLUX model in terms of text alignment.
Moreover, our method also enhances adherence to the input condition across various baselines. For instance, on the MSE loss, our approach reduces the error of FLUX-Depth by \textbf{148.687}, demonstrating improved conditional fidelity.
Finally, Tab.~\ref{tab:coco_unionpro2_depth_canny} shows that our approach does not compromise the base model’s robustness: on COCO—a dataset not used during training—it still delivers improvements over the original model. Importantly, all results shown here are obtained without iterative optimization.

\begin{table}[t]
\centering
\caption{\rebuttal{\textbf{Quantitative results on COCO Benchmark with depth and canny conditioning.}}}
\vspace{-4mm}
\footnotesize
\setlength{\tabcolsep}{1.8mm}
\begin{tabular}{l | cccc | cccc}
\toprule[1pt]
 & \multicolumn{4}{c|}{\textbf{Depth-conditioned}} & \multicolumn{4}{c}{\textbf{Canny-conditioned}} \\
\cmidrule(lr){2-5} \cmidrule(lr){6-9}
\textbf{Method} & \textbf{SR} $\uparrow$ & \textbf{MSE} $\downarrow$ & \textbf{SGMSE} $\downarrow$ & \textbf{CLIP} $\uparrow$
                & \textbf{SR} $\uparrow$ & \textbf{F1} $\uparrow$ & \textbf{SGF1} $\uparrow$ & \textbf{CLIP} $\uparrow$ \\
\midrule

\rowcolor[HTML]{F5F5F5}
\rebuttal{LooseControl} & \rebuttal{0.72} & \rebuttal{1334.0} & \rebuttal{1706.0} & \rebuttal{0.2534} & \rebuttal{N/A} & \rebuttal{N/A} & \rebuttal{N/A} & \rebuttal{N/A} \\

\rowcolor[HTML]{F5F5F5}
\rebuttal{ControlNet++} & \rebuttal{0.79} & \rebuttal{548.3} & \rebuttal{668.9} & \rebuttal{0.2557} & \rebuttal{0.71} & \rebuttal{0.339} & \rebuttal{0.245} & \rebuttal{0.2622} \\

\hline

\rowcolor[HTML]{F8FBFF}
Union-Pro2      & 0.83 & 297.3 & 363.5 & 0.2546 & 0.78 & 0.416 & 0.332 & 0.2539 \\
\rowcolor[HTML]{F8FBFF}
+ SFT           & 0.87 & 561.7 & 635.6 & 0.2575 & 0.81 & 0.271 & 0.271 & 0.2602 \\
\rowcolor[HTML]{F8FBFF}
+ DPO           & 0.90 & 263.4 & 278.2 & 0.2586 & 0.75 & 0.490 & 0.373 & 0.2554 \\
\rowcolor[HTML]{F8FBFF}
+ Ours          & \textbf{0.91} & \textbf{236.3} & \textbf{245.3} & \textbf{0.2633} & \textbf{0.83} & \textbf{0.497} & \textbf{0.392} & \textbf{0.2629} \\
\bottomrule[1pt]
\end{tabular}
\label{tab:coco_unionpro2_depth_canny}
\end{table}

\begin{figure}[t!]
	\includegraphics[width=1.0\linewidth]{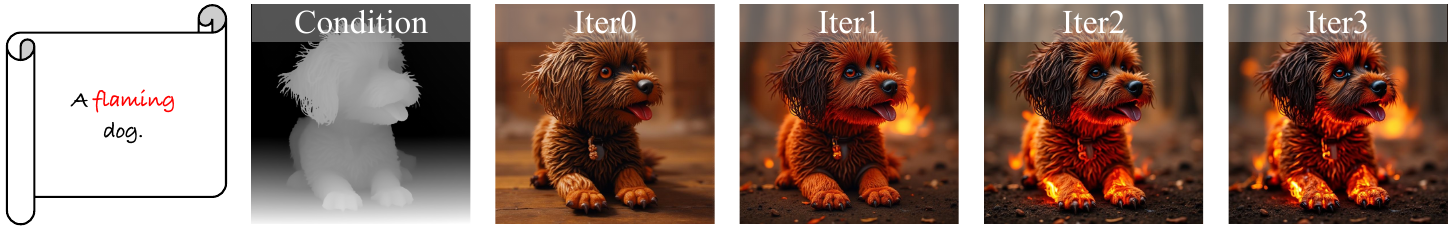}
	\vspace{-6mm}
	\caption{
		\textbf{Visualization of Iterative Optimization.}
	}
	\vspace{-6mm}
	\label{fig:iter_vis}
\end{figure}

\begin{wraptable}[12]{r}{0.55\linewidth} 
\centering
    \vspace{-0.5cm}
\caption{\textbf{Ablation study of our core components.} ``w/o ALB'' ablates our Adaptive Loss Balancing. ``Text. Only'' and ``Cond. Only'' are trained using only the text or condition preference pairs, respectively.}
\vspace{-4mm}
\footnotesize
\setlength{\tabcolsep}{1.7mm}%
{
	\centering
	\begin{tabular}{lcccc}
        \bottomrule
		\textbf{Method}   & \textbf{SR} $\uparrow$ & \textbf{MSE} $\downarrow$ & \textbf{SGMSE} $\downarrow$ & \textbf{CLIP} $\uparrow$ \\
		\hline
		\rowcolor[HTML]{F8FBFF}
		\rowcolor[HTML]{F8FBFF}
            Iter = 1                & 0.84
                              & 163.968
                         & 195.728
                           & 0.2924

                           \\
		\rowcolor[HTML]{F8FBFF}
            Iter = 2                & 0.85
                              & \textbf{158.876}
                         & \underline{195.459}
                           & \underline{0.2947}

                           \\
		\rowcolor[HTML]{F8FBFF}
            \textbf{Iter = 3 (Ours)}               & \textbf{0.88}
                              & \underline{159.559}
                         & \textbf{190.263}
                           & \textbf{0.2957}

                           \\
		\rowcolor[HTML]{F8FBFF}
            Iter = 4                & \underline{0.86}
                             & 166.274
                         & 202.363
                          & 0.2939

                           \\
                           \hline
		\rowcolor[HTML]{F5F5F5}
		w/o ALB                & 0.78                             & {157.729}                         & 205.246                          & 0.2862
                           \\
        \rowcolor[HTML]{F5F5F5}
		Text. Only                &  0.88                            & 258.954                         & 287.749                          & 0.2947
                           \\
        \rowcolor[HTML]{F5F5F5}
            Cond. Only               & 0.59                             & 153.659                       & 218.683                         & 0.2753
                           \\
		\toprule
	\end{tabular}
	\vspace{-4mm}
    \label{tab:iter_results}
}
\end{wraptable}

\noindent\textbf{Iterative optimization strategy.}
Since BideDPO simultaneously strengthens the model’s adherence to both text and condition, we can adopt an iterative optimization algorithm to refine the model and data together.  
Tab.~\ref{tab:iter_results} shows that our iterative optimization (Fig.~\ref{fig:overview}) progressively improves adherence to both condition and text, reaching optimal performance by the third iteration. This trend is further supported by the qualitative comparisons in Fig.~\ref{fig:iter_vis}, where generated images increasingly align with the text while preserving condition fidelity. Importantly, even a single iteration already yields substantial improvements over the baseline. Thus, iterative refinement should be regarded as an optional enhancement that can be adjusted based on available computational resources.

\noindent\textbf{Ablation Study of Adaptive Loss Balancing (ALB).}
In Tab.~\ref{tab:iter_results}, the results for “Iter = 1” demonstrate more balanced improvements than those for “w/o ALB”, with a \textbf{6\%} increase in success rate and a \textbf{9.518} gain in SGMSE, despite only a modest decrease in MSE (\textbf{6.239}).

\noindent\textbf{Ablation Study on Preference Pairs.}
As shown in Tab.~\ref{tab:iter_results}, using only text- or condition-disentangled preference pairs (Fig.~\ref{fig:data_pipe}) makes the model focus on a single aspect, yielding marginal or even harmful effects on the other. 
Only by jointly leveraging both types of preference pairs (ours) can the model achieve balanced improvements across all constraints.

\noindent\textbf{User Study.}
We conducted a user study to compare our optimized model against the base model, evaluating three aspects: text adherence, condition adherence, and overall alignment. 
For each trial, we randomly sampled 20 cases and asked 30 participants to evaluate them in a 1-\textit{vs.}-1 format, yielding a total of 600 comparisons. As shown in Tab.~\ref{tab:user_study}, participants favored our model roughly twice as often as the baseline.

\begin{table}[t]
\centering

\begin{minipage}[t]{0.5\linewidth} 
\centering

\captionof{table}{\textbf{User study results.} Values are win rates.}
\vspace{-3mm} 
\label{tab:user_study}
\footnotesize
\setlength{\tabcolsep}{0mm} 

\begin{tabular*}{\linewidth}{l@{\extracolsep{\fill}}ccc}
\toprule[1pt]
\textbf{Comparison} & \textbf{Text} $\uparrow$ & \textbf{Condition} $\uparrow$ & \textbf{Overall} $\uparrow$ \\
\midrule
Ours \textit{vs.} Base & 67.9\% & 66.8\% & 64.0\% \\
\bottomrule[1pt]
\end{tabular*}

\vspace{-3mm} 

\captionof{table}{\rebuttal{\textbf{Style-conditioned generation.}}}
\label{tab:style_results}
\footnotesize
\setlength{\tabcolsep}{0mm}

\begin{tabular*}{\linewidth}{l@{\extracolsep{\fill}}cccc}
\toprule[1pt]
\textbf{Method} & \textbf{SR} $\uparrow$ & \textbf{Style} $\uparrow$ & \textbf{SG Style} $\uparrow$ & \textbf{CLIP} $\uparrow$ \\
\midrule
IPA & 30\% & \textbf{6.50} & 1.31 & 0.1679 \\
+ Ours & \textbf{58\%} & 6.26 & \textbf{2.97} & \textbf{0.2015} \\
\bottomrule[1pt]
\end{tabular*}
\end{minipage}
\hfill
\begin{minipage}[t]{0.48\linewidth}
\centering
\captionof{table}{\rebuttal{\textbf{Success rates from different evaluators.} Results show that our improvements are not artifacts of a specific VLM but hold universally across models and human judge.}}
\vspace{-2.3mm}
\label{tab:vlm_validation_main_paper}
\footnotesize
\setlength{\tabcolsep}{1.2mm} 
\begin{tabular}{lcccc}
\toprule[1pt]
Judge & UnionPro2 & +SFT & +DPO & +Ours \\
\midrule
Qwen2.5 & 0.49 & 0.70 & 0.71 & \textbf{0.84} \\
GPT-4o & 0.43 & 0.65 & 0.70 & \textbf{0.82} \\
Human & 0.42 & 0.62 & 0.64 & \textbf{0.84} \\
\bottomrule[1pt]
\end{tabular}
\end{minipage}
\vspace{-2mm}
\end{table}

\begin{figure*}[tb!]
	\includegraphics[width=1.0\linewidth]{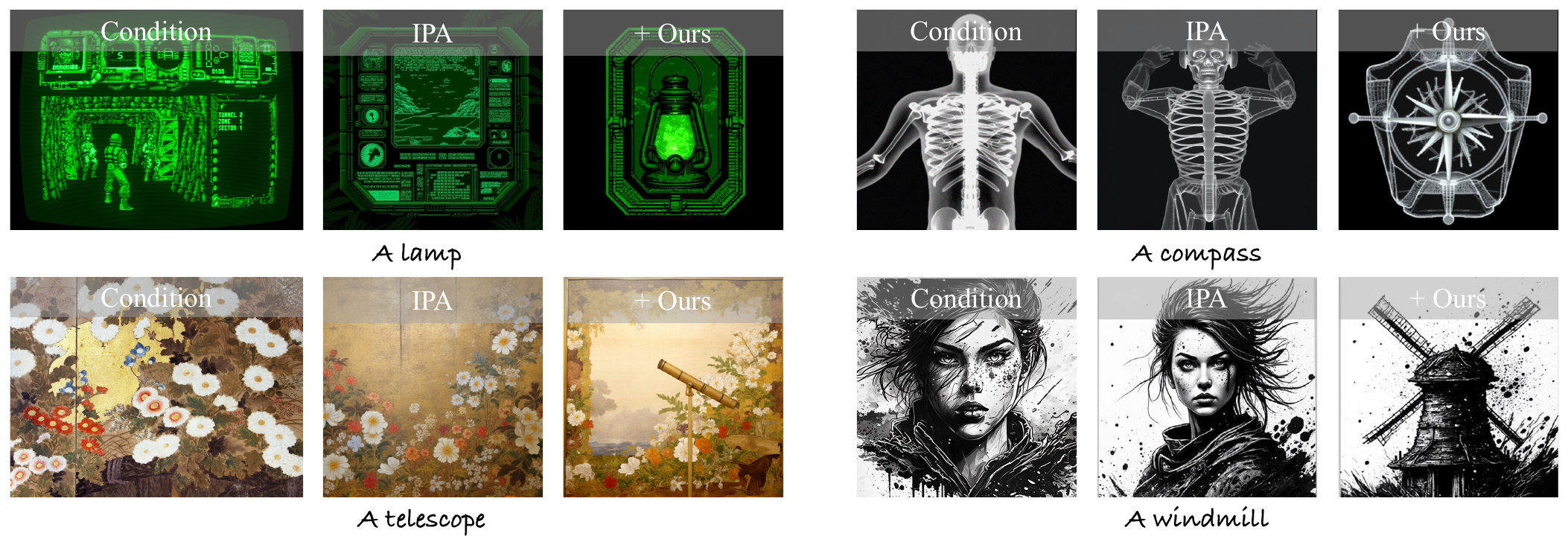}
	\vspace{-5mm}
	\caption{
		\rebuttal{\textbf{Visual results for style-conditioned image generation on IP-Adapter~\citep{flux-ipa}.}}
	}
	\label{fig:style_vis_main_paper}
    \vspace{-5mm}
\end{figure*}

{\noindent\textbf{Universality on style-conditional image generation.}}
{Beyond structure- and spatial-conditional generation, we further validate the effectiveness of our method on more abstract conditions, such as style-conditioned image generation. As shown in Fig.~\ref{fig:style_vis_main_paper}, prior methods often suffer from excessive reference copying, due to the Model Bias issue discussed in Fig.~\ref{fig:effect_vis}(b). In contrast, our proposed BideDPO algorithm substantially mitigates this problem, achieving a 28\% higher success rate (Tab.~\ref{tab:style_results}).}

\rebuttal{\noindent\textbf{More VLM Selection.}}
\label{sec:discussion_vlm}
\rebuttal{To verify robustness beyond a single evaluator, we additionally assessed our method using multiple VLMs---including Qwen2.5-VL-72B, GPT-4o---and human raters. As shown in Tab.~\ref{tab:vlm_validation_main_paper}, all evaluators yield consistent rankings, confirming that our improvements are not artifacts of a specific VLM but hold universally across models and human judgment.}
\section{Conclusion}

In this work, we address the fundamental challenge of achieving simultaneous text and condition alignment in controllable image generation. We identify that existing approaches—including supervised fine-tuning and naive DPO—struggle to balance multiple constraints, especially when the text prompt and condition input are in conflict. To overcome this, we propose a bidirectionally decoupled DPO framework that disentangles the optimization of textual and conditional adherence. 
Furthermore, adaptive loss balancing ensures stable and effective learning between objectives.
Our approach also features an automated pipeline for constructing high-quality, disentangled preference pairs, as well as an iterative optimization strategy that continuously enhances both the data and the model. 
Extensive experiments demonstrate that our method significantly outperforms strong baselines on both textual and conditional alignment, yielding substantial improvements in Success Ratio and conditional fidelity across a variety of benchmarks. Our framework not only advances the state of controllable image generation, but also provides new insights into preference-based learning with multiple, potentially conflicting objectives.
\clearpage

\bibliography{2026_conference}
\bibliographystyle{2026_conference}

\clearpage
\appendix
\section{Preliminary}
\label{sec:preliminary}

\noindent
\textbf{Denoising Diffusion Models.}
Denoising diffusion models~\cite{ddpm} are a class of generative models that learn to synthesize data by reversing a fixed forward noising process. This process gradually adds Gaussian noise to a clean sample $x_0$ over $T$ timesteps, such that $x_t = \sqrt{\bar{\alpha}_t}x_0 + \sqrt{1-\bar{\alpha}_t}\epsilon$, where $\epsilon \sim \mathcal{N}(0, \mathbf{I})$. A neural network $\epsilon_\theta$ is then trained to predict the added noise $\epsilon$ from the noisy sample $x_t$ timestep $t$, and context $c$. The objective is to minimize the L2 error between the actual and predicted noise:

By learning to effectively denoise at every step, the model can generate high-fidelity samples from pure noise.
{
\begin{equation}
    \mathcal{L}_{\text{simple}} = \mathbb{E}_{\epsilon, t, x_0, x_t}\left[ \lVert \epsilon - \epsilon_\theta(x_t, t, c) \rVert_2^2 \right].
\end{equation}
}%
\noindent\textbf{Direct Preference Optimization (DPO).} 
\citet{directpreferenceoptimization} introduced DPO, a method to fine-tune LLMs with pairs of ranked examples $(x^+, x^-, c)$, where $x^+$ is the preferred and $x^-$ the dispreferred sample. The training objective is formulated using an implicit reward $\hat{r}_{\theta}(x,c)=\beta\cdot\log{\frac{p_\theta(x|c)}{p_{\mbox{\tiny{ref}}}(x|c)}}$, which measures the log-likelihood ratio with respect to a reference model $p_{\mbox{\scriptsize{ref}}}$ and $\beta$ is a hyperparameter. The DPO loss is expressed as:
{
\begin{equation}
	\label{eq_dpo}
	\begin{split}
		 \mathcal{L}_\mathrm{\mbox{\scriptsize{DPO}}}(\theta) = -\mathbb{E}_{x^+, x^-, c} \left[\log\sigma\left( \hat{r}_{\theta}(x^+, c) - \hat{r}_{\theta}(x^-, c) \right)\right] \\
		 = -\mathbb{E}_{x^+\!, x^-, c} \!\left[\!\log\!\sigma\!\bigg(\!\!\beta\!
			\left(\!\!\log\!{\frac{p_\theta(x^+\!|c)}{p_{\mbox{\scriptsize{ref}}}(x^+\!|c)}}\!-\!\log\!{\frac{p_\theta(x^-\!|c)}{p_{\mbox{\scriptsize{ref}}}(x^-\!|c)}}\!\!\right)\!\!\!\!\bigg)\!\!\right]\!\!,
	\end{split}
\end{equation}
}%
where $\sigma$ is the sigmoid function. 

\noindent
\textbf{Diffusion-DPO}. \citet{diffusiondpo} applied DPO to diffusion models by modifying Eq.~\ref{eq_dpo}. They replaced the logarithmic difference by the denoising error:
\begin{equation}
	\label{eq_diffusion_dpo}
	\begin{split}
		\!\mathcal{L}_{\mathrm{\mbox{\scriptsize{D-DPO}}}} & (\theta) \!=\! - \mathbb{E}_{x_t^+, x_t^-, c}\!\Big[\!\log\sigma\Big(\!\!-\!\beta T 
		\Big(                                                                                                                                             \\ &
			\lVert\epsilon^+\!-\!\epsilon_\theta^+(x_t^+\!, t, c)\rVert_2^2\!-\!
		\lVert\epsilon^+\!-\!\epsilon_{\mbox{\scriptsize{ref}}}^+(x_t^+, t, c)\rVert_2^2 -                                                                \\ &
			\lVert\epsilon^-\!-\!\epsilon_\theta^-(x_t^-, t, c)\rVert_2^2\!+\!
			\lVert\epsilon^-\!-\!\epsilon_{\mbox{\scriptsize{ref}}}^-(x_t^-, t, c)\rVert_2^2
			\!\Big)\!\Big)\!\Big]\!,
	\end{split}
\end{equation}
where $x_t^+$ and $x_t^-$ are obtained  from $x_0^+$ and $x_0^-$ using the forward process of diffusion models, and $T$ is a temperature.

\section{Implementation Details}

\subsection{Baseline Models}
All our experiments are conducted using state-of-the-art conditional text-to-image diffusion models from the FLUX family~\citep{flux}. Specifically, we adopt the following publicly available, pre-trained models as our baselines:
\begin{itemize}
	\item \textbf{FLUX-Depth:} A variant specialized for depth-conditioned image generation, implemented by the FLUX team.
	\item \textbf{FLUX-Canny:} A variant specialized for Canny-edge-conditioned image generation, implemented by the FLUX team.
	\item \textbf{Union-Pro2}~\citep{unionpro2}: A powerful model built on top of FLUX that supports conditional generation. It incorporates the ControlNet approach by adding an extra side network to the FLUX architecture, enabling multi-modal conditioning. Union-Pro2 is one of the most widely downloaded models in the AIGC community.
	\item \rebuttal{\textbf{FLUX.1-dev-IP-Adapter}~\citep{flux-ipa}: An IP-Adapter implementation for FLUX.1-dev released by InstantX Team, enabling image-conditioned generation where reference images provide style conditions for generation.}
\end{itemize}
We primarily compare our proposed BideDPO method against two standard baselines: Supervised Fine-Tuning (SFT) and a naive application of DPO~\citep {diffusiondpo}, evaluated on the models listed above.

\subsection{Baseline Configurations}
To ensure a fair comparison, we configured the Supervised Fine-Tuning (SFT) and naive DPO baselines as follows, using the preference data generated by our pipeline:

\textbf{SFT.} The SFT baseline was trained with a standard denoising score matching objective. We exclusively used high-quality positive samples from our generated data—specifically, the preferred samples $x_T^+$ from the text-disentangled pairs (conditioned on the target prompt $p$ and initial condition $s_0$) and $x_C^+$ from the condition-disentangled pairs (conditioned on $p$ and the refined condition $s_1$).

\textbf{Naive DPO.} For the DPO baseline, we constructed text-and-condition preference pairs to simulate a standard DPO setting in which preferences are not disentangled:
\begin{itemize}
  \item The preferred sample $x^{+}$ is $x_C^+$ from the condition-disentangled pair, which aligns well with both the target prompt $p$ and the condition $s_1$.
  \item The dispreferred sample $x^{-}$ is $x_T^-$ from the text-disentangled pair. This sample is generated from the initial condition $s_0$ and does not align with the target prompt $p$, making it a poor fit for both the target prompt $p$ and the condition $s_1$.
\end{itemize}
In this way, $x_C^+$ and $x_T^-$ together form a DPO preference pair that jointly enforces adherence to both the target prompt $p$ and the condition $s_1$.

\subsection{Training Hyperparameters}
All models were fine-tuned on a cluster of 4 NVIDIA A100 GPUs with a batch size of 4. We employed the Low-Rank Adaptation (LoRA)~\citep{hu2022lora} method with a rank of 256 for fine-tuning all models. For the SFT method, we trained for 5,000 steps using the Prodigy~\citep{mishchenko2023prodigy} optimizer with a learning rate of 1.0. For both DPO and BideDPO, starting from the SFT-tuned model, we further optimized for an additional 2,000 steps using the AdamW~\citep{adam} optimizer with a learning rate of 0.00004 and a weight decay of 0.01. We set the $\beta$ parameter~\citep{diffusiondpo} for DPO to 5000. \rebuttal{Each iteration (data generation + fine-tuning) uses 8×A800 GPUs (40GB) for 5 hours (data generation) and 4×A100 GPUs (80GB) for 3 hours (training).}

\section{Evaluation Metric Details}
In our experiments, we employ a comprehensive set of metrics to evaluate model performance across text alignment, conditional fidelity, and their combination. Below, we provide detailed descriptions of each metric.

\subsection{Text Alignment Metrics}

\subsubsection{Success Ratio}
To automatically assess whether a generated image accurately reflects the text prompt, we use the powerful Vision-Language Model (VLM) \texttt{Qwen2.5-VL-72B}. For each generated image and its corresponding target prompt, we query the VLM with a carefully designed question: ``Does the image successfully depict the following description: `[Prompt]'? Please answer with `Yes' or `No'.'' The Success Ratio is then calculated as the percentage of ``Yes'' responses across our entire test set. This provides a direct and automated measure of text-image consistency.

\subsubsection{CLIP Score}
The CLIP Score measures the semantic similarity between the generated image and the input text prompt. We use the pre-trained ViT-L/14 CLIP model to compute embeddings for both the image and the prompt. The score is the cosine similarity between these two embedding vectors, scaled by 100. A higher CLIP score indicates better semantic alignment with the user's textual description.

\subsection{Conditional Fidelity Metrics}

\subsubsection{Mean Squared Error (MSE) and F1 Score}
To quantify how well a generated image adheres to the input structural condition (\textit{e.g.}, depth map, Canny edges), we first extract the corresponding condition map from the generated image using the same tool employed during data creation (\textit{e.g.}, Depth Anything v2 for depth). We then compute the Mean Squared Error (MSE) or F1 Score between the extracted condition map and the original input condition map.
\begin{itemize}
	\item \textbf{MSE:} Used for pixel-wise regression tasks like depth map prediction. A lower MSE indicates higher fidelity to the ground-truth condition.
	\item \textbf{F1 Score:} Used for tasks like Canny edge or human pose matching, where we can treat it as a binary segmentation problem. A higher F1 score indicates better structural correspondence.
\end{itemize}

\subsection{Combined Text and Condition Metrics}

\subsubsection{Semantic-Guided MSE (SGMSE) and F1 (SGF1)}
Standard conditional metrics like MSE and F1 only measure structural fidelity and ignore whether the generated image is semantically correct according to the text prompt. To address this, we introduce two novel metrics: Semantic-Guided MSE (SGMSE) and Semantic-Guided F1 (SGF1). These metrics integrate a semantic check (using the same VLM as for the Success Ratio) into the calculation:
\begin{itemize}
\item If the generated image is deemed a ``Success'' (\textit{i.e.}, it matches the text prompt), the SGMSE and SGF1 are the same as the standard MSE and F1 scores.
\item If the generated image is a ``Failure'' (it does not match the text prompt), we apply a penalty to reflect the semantic mismatch. The SGMSE is doubled (\textit{i.e.}, $2 \times \text{MSE}$), and the SGF1 score is set to zero.
\end{itemize}
This penalty mechanism ensures that the model is rewarded only when it satisfies both the textual and conditional constraints simultaneously, providing a more holistic evaluation of conditional generation.

\section{Data Construction Details}
\subsection{Condition Modalities}
Our framework is designed to be agnostic to the specific type of conditional input. For the experiments in this paper, we constructed datasets for three different structural modalities:
\begin{itemize}
	\item \textbf{Depth Maps:} To provide 3D scene geometry, we utilized the widely-used \texttt{Depth Anything v2}~\citep{yang2024depthv2} model to extract high-quality depth maps from images.
	\item \textbf{Canny Edges:} For sharp, well-defined object boundaries, we used the standard Canny edge detection algorithm.
	\item \textbf{Soft Edges:} We employ the {\em ControlNet-SoftEdge} family of edge detectors, specifically the recent \texttt{MistoLine-SDXL} model~\citep{softedge}. 
\end{itemize}
This variety of conditions allows us to evaluate the robustness and versatility of our method across different types of structural constraints.

\subsection{More Details of Automated Data Construction Pipeline}
Our data construction pipeline is designed to automatically generate disentangled preference pairs for both text and condition alignment. This process is crucial for training our model to handle multi-objective optimization effectively, especially in cases with conflicting constraints. The pipeline, as illustrated in the main paper, consists of the following three steps:

\subsubsection{Step 1: Prompt and Initial Condition Generation}
The process begins with the generation of prompts using a large language model (LLM). For each data point, we generate a basic \textit{Source Prompt} and a more descriptive \textit{Target Prompt} ($p$). The source prompt is then used to create an initial, often loose, condition map, which we denote as ``Condition 0'' ($s_0$). This initial pairing of the target prompt $p$ and Condition 0 $s_0$ is intentionally designed to often contain conflicts, either at the input level or due to model priors, as discussed in the main paper.

\subsubsection{Step 2: Text-Disentangled Pair Generation}
With the prompts and initial condition, we generate the text-disentangled preference pair $(x_T^+, x_T^-, p, s_{0})$. Both samples in this pair are generated to adhere to the same initial ``Condition 0'' ($s_0$).
\begin{itemize}
	\item \textbf{Preferred Sample ($x_T^+$):} The preferred sample is generated using the detailed \textit{Target Prompt}. Its alignment with the text is verified using a Vision-Language Model (VLM). This high-quality sample serves as a reference anchor, denoted $x_a$.
	\item \textbf{Dispreferred Sample ($x_T^-$):} The dispreferred sample is generated using the basic \textit{Source Prompt}. As a result, it correctly follows ``Condition 0'' but lacks the specific textual details present in the target prompt, making it less preferred from a text-alignment perspective.
\end{itemize}

\subsubsection{Step 3: Condition-Disentangled Pair Generation}
Next, we construct the condition-disentangled preference pair $(x_C^+, x_C^-, p, s_{1})$. For this pair, both samples are generated to align with the same \textit{Target Prompt} $p$.
\begin{itemize}
	\item \textbf{Preferred Sample ($x_C^+$):} The anchor image $x_a$ from the previous step is used as the preferred sample. A new, strictly aligned condition map, ``Condition 1'' ($s_1$), is then extracted directly from this anchor image.
	\item \textbf{Dispreferred Sample ($x_C^-$):} The dispreferred sample is generated to match the semantics of the target prompt but to adhere less strictly to the new ``Condition 1''. This creates a preference based on conditional fidelity. Because the generator has limited capability, the generated image $x_C^-$ will inevitably exhibit some loss of fidelity to the precise structural details of $s_1$ when compared to the original image $x_a$ from which $s_1$ was derived.
\end{itemize}

This structured, three-step process allows us to systematically generate a large dataset of preference pairs that isolate and target distinct aspects of text and condition alignment.

\rebuttal{\section{BideDPO on style-conditioned Generation}}

\begin{figure*}[tb!]
	\includegraphics[width=1.0\linewidth]{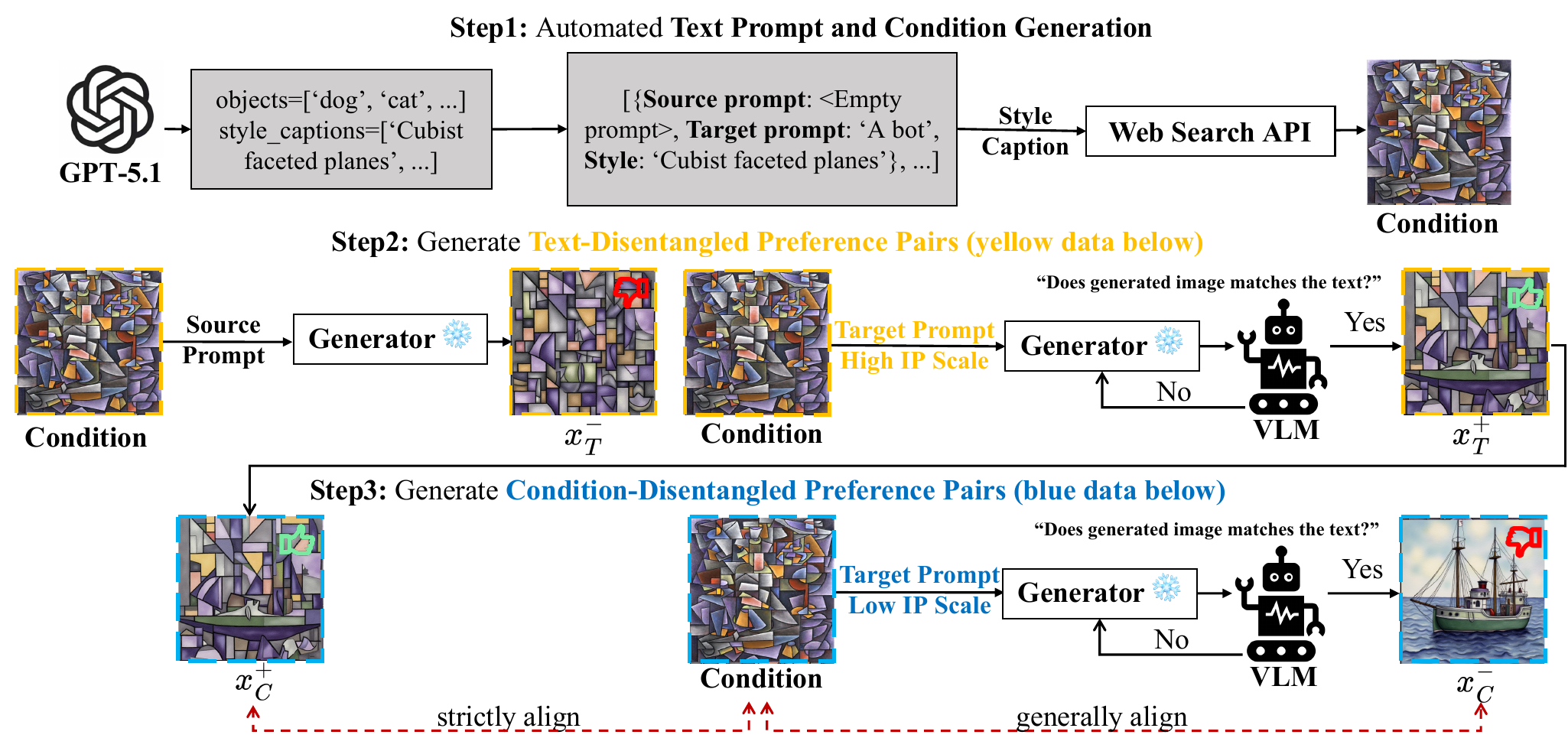}
	\vspace{-6mm}
	\caption{
		\rebuttal{\textbf{The Automated Style-Aware Preference Data Pipeline.} }
	}
	\vspace{-4mm}
	\label{fig:data_style_pipe}
\end{figure*}

\rebuttal{\subsection{{Automated Style-Aware Preference Pipeline}}}
\rebuttal{To incorporate explicit artistic controls into our preference data, we extend the above pipeline with the style-aware branch illustrated in Fig.~\ref{fig:data_style_pipe}. As shown in Step~1, GPT-5.1 enumerates object concepts together with concise style captions (``Cubist faceted planes,'' etc.). The LLM emits a minimalist \textit{Source Prompt}, a descriptive \textit{Target Prompt}, and the style caption, which is forwarded to a Web-Search API to retrieve a representative condition image.}

\rebuttal{Step~2 mirrors the step 2 of Fig.~\ref{fig:data_pipe}. Holding the retrieved condition fixed, we render a positive sample $x_{T,\text{style}}^+$ using the Target Prompt and a high IP-Adapter scale, with VLM verification ensuring that both the textual semantics and the referenced style are expressed. A negative counterpart $x_{T,\text{style}}^-$ is produced with the minimalist Source Prompt (empty string) while reusing the same condition, yielding an image that resembles the structure and style but fails to mention the target concept. A VLM adversary inspects each pair, only accepting anchors for which the text truly matches the prompt and the style hint.}

\rebuttal{Step~3 then isolates conditional fidelity. We keep the Target Prompt and style condition fixed but vary the IP-Adapter scale so that $x_{C,\text{style}}^+$ strictly follows the retrieved condition map while $x_{C,\text{style}}^-$ only coarsely aligns (``strictly align'' vs.\ ``generally align'' in the figure). Both samples still satisfy the textual description, so their difference arises purely from how faithfully they respect the style-conditioned control input. This yields preference pairs $(x_{C,\text{style}}^+, x_{C,\text{style}}^-, p, s_1)$ that drop seamlessly into the unified BideDPO training mix. Notably, this style-aware pipeline follows almost the same structure as our main data pipeline, highlighting the generality and versatility of our proposed preference data construction framework.}

\rebuttal{\subsection{Style Benchmark and Data.}
\label{sec:style_benchmark}
Using the data generation pipeline defined in the Automated Style-Aware Preference Pipeline above, we construct a style-conditioned benchmark. The training set consists of 100 objects and 20 styles, while the test set contains 10 different objects and 10 different styles.}

\rebuttal{\textbf{Metrics.}~~We ask Qwen2.5-VL 72B~\cite{bai2023qwen} to rate each generated image on a 0--10 \textbf{Style Score}, and define \textbf{SG Style Score} by zeroing the rating whenever the reference style conflicts with the Target Prompt. We also track success rate (SR)---the fraction of samples where the VLM judges both text and style as satisfied---along with CLIP similarity for semantic grounding. These metrics jointly capture textual fidelity, structural accuracy, and adherence to the curated style references.}

\rebuttal{\subsection{{Experiment Results}}}

\rebuttal{\textbf{Quantitative Results.}~~Tab.~\ref{tab:style_results} presents the comprehensive quantitative comparison on the DualAlign-Style benchmark. As shown, BideDPO significantly improves the IP-Adapter (IPA) baseline across all key metrics. Most notably, the success rate (SR) increases from \textbf{30\% to 58\%}, representing a \textbf{93\% relative improvement}, which demonstrates that our method substantially enhances the model's ability to simultaneously satisfy both textual semantics and style constraints. While the baseline achieves a slightly higher mean Style Score (6.50 vs. 6.26), this is expected because IPA tends to directly copy the style reference image, which naturally yields high style similarity scores but at the cost of semantic accuracy (only 30\% SR). In contrast, the disentangled \textbf{SG Style Score} reveals the true capability: BideDPO achieves \textbf{2.97} compared to IPA's \textbf{1.31}, representing a \textbf{127\% relative improvement}. This substantial gap demonstrates that BideDPO effectively resolves conflicts between style references and text prompts, producing outputs that maintain style fidelity even when the style condition is semantically incompatible with the target description, while IPA's high Style Score primarily reflects its tendency to copy the reference rather than harmonize style with semantics. Additionally, the CLIP score improves from 0.1679 to 0.2015, confirming better semantic alignment with the text prompts. These results demonstrate that the same bidirectionally decoupled objective extends seamlessly to abstract, reference-image style conditions without requiring architectural modifications, simply by adding another preference head and an adaptive loss balancer weight.}

\rebuttal{\textbf{Qualitative Results.}~~Fig.~\ref{fig:style_vis_main_paper} presents comprehensive qualitative comparisons on the DualAlign-Style benchmark, showcasing BideDPO's superior capability in style-conditioned generation. The visualization demonstrates four challenging generation tasks, each requiring the model to generate a specific target object (a lamp, a compass, a telescope, and a windmill) while simultaneously adhering to diverse and complex style conditions. These conditions span a wide range of artistic styles: green pixelated retro game aesthetics, X-ray transparency effects, traditional Japanese art with gold leaf accents, and gritty ink-splatter illustrations. As shown in the figure, the IP-Adapter (IPA) baseline often fails to generate the specified object, instead producing variations of the condition itself or semantically related but incorrect content. For instance, when asked to generate ``A lamp'' under a retro game style condition, IPA produces a green-tinted futuristic UI overlay without the target lamp object. Similarly, for ``A compass'' under an X-ray condition, IPA generates a human skeleton instead of the requested compass. In contrast, our BideDPO method successfully integrates the target object into the given condition's aesthetic, demonstrating superior capability in harmonizing textual semantics with stylistic constraints while maintaining both object accuracy and style fidelity. The generated images not only correctly depict the target objects but also faithfully preserve the distinctive visual characteristics of each style condition, such as the green monochrome digital aesthetic for the lamp, the transparent wireframe style for the compass, the ornate floral background with gold accents for the telescope, and the gritty ink-splatter texture for the windmill.}

\begin{algorithm}[h!]
\caption{Automated Construction of Disentangled Preference Data}
\label{alg:data_construction}
\begin{algorithmic}[1]
\STATE \textbf{Input:} Generator model $G$, LLM, Condition extractor $E$, VLM, Num samples to generate $N$, Max retries $K$.
\STATE \textbf{Initialize:} Unified preference set $\mathcal{D} \leftarrow \emptyset$.

\STATE \textit{Step 1: Pre-generate prompts and initial conditions}
\STATE $P_{\text{pool}} \leftarrow \text{LLM.generate\_source\_target\_pairs()}$
\STATE $\text{Context}_{\text{pool}} \leftarrow \emptyset$
\FOR{$(p_{\text{source}}, p_{\text{target}})$ in $P_{\text{pool}}$}
    \STATE $x_{\text{init}} \leftarrow G(p_{\text{source}})$
    \STATE $s_0 \leftarrow E(x_{\text{init}})$
    \STATE $\text{Context}_{\text{pool}} \leftarrow \text{Context}_{\text{pool}} \cup \{(p_{\text{source}}, p_{\text{target}}, s_0)\}$
\ENDFOR

\WHILE{$|\mathcal{D}| < N$}
    \STATE $p_{\text{source}}, p_{\text{target}}, s_0 \leftarrow \text{RandomSample}(\text{Context}_{\text{pool}})$
    
    \STATE \textit{--- Step 2: Attempt to generate text-disentangled pair ---}
    \STATE $x_a \leftarrow \text{None}$
    \STATE $tries \leftarrow 0$
    \WHILE{$tries < K$}
        \STATE candidate $\leftarrow G(p_{\text{target}}, s_0)$
        \IF{VLM.verify(candidate, $p_{\text{target}}$)}
            \STATE $x_a \leftarrow \text{candidate}$
            \STATE \textbf{break}
        \ENDIF
        \STATE $tries \leftarrow tries + 1$
    \ENDWHILE
    \IF{$x_a$ is \text{None}} \STATE \textbf{continue} \COMMENT{Failed to generate a valid anchor} \ENDIF
    
    \STATE $x_T^+ \leftarrow x_a$
    \STATE $x_T^- \leftarrow G(p_{\text{source}}, s_0)$
    
    \STATE \textit{--- Step 3: Attempt to generate condition-disentangled pair ---}
    \STATE $x_C^+ \leftarrow x_a$
    \STATE $s_1 \leftarrow E(x_a)$
    \STATE $x_C^- \leftarrow \text{None}$
    \STATE $tries \leftarrow 0$
    \WHILE{$tries < K$}
        \STATE candidate $\leftarrow G(p_{\text{target}}, s_1)$
        \IF{VLM.verify(candidate, $p_{\text{target}}$)}
            \STATE $x_C^- \leftarrow \text{candidate}$
            \STATE \textbf{break}
        \ENDIF
        \STATE $tries \leftarrow tries + 1$
    \ENDWHILE
    \IF{$x_C^-$ is \text{None}} \STATE \textbf{continue} \COMMENT{Failed to generate a valid counterpart} \ENDIF

    \STATE \textit{--- Both pairs successfully generated, add to unified dataset ---}
    \STATE $c_0 \leftarrow (p_{\text{target}}, s_0)$
    \STATE $c_1 \leftarrow (p_{\text{target}}, s_1)$
    \STATE $pair_T \leftarrow (x_T^+, x_T^-, c_0)$
    \STATE $pair_C \leftarrow (x_C^+, x_C^-, c_1)$
    \STATE $\mathcal{D} \leftarrow \mathcal{D} \cup \{ (pair_T, pair_C) \}$
\ENDWHILE
\STATE \textbf{return} $\mathcal{D}$
\end{algorithmic}
\end{algorithm}

\section{Additional Visualization Results}
\label{sec:additional_vis}
\begin{figure*}[ht!]
	\includegraphics[width=1.0\linewidth]{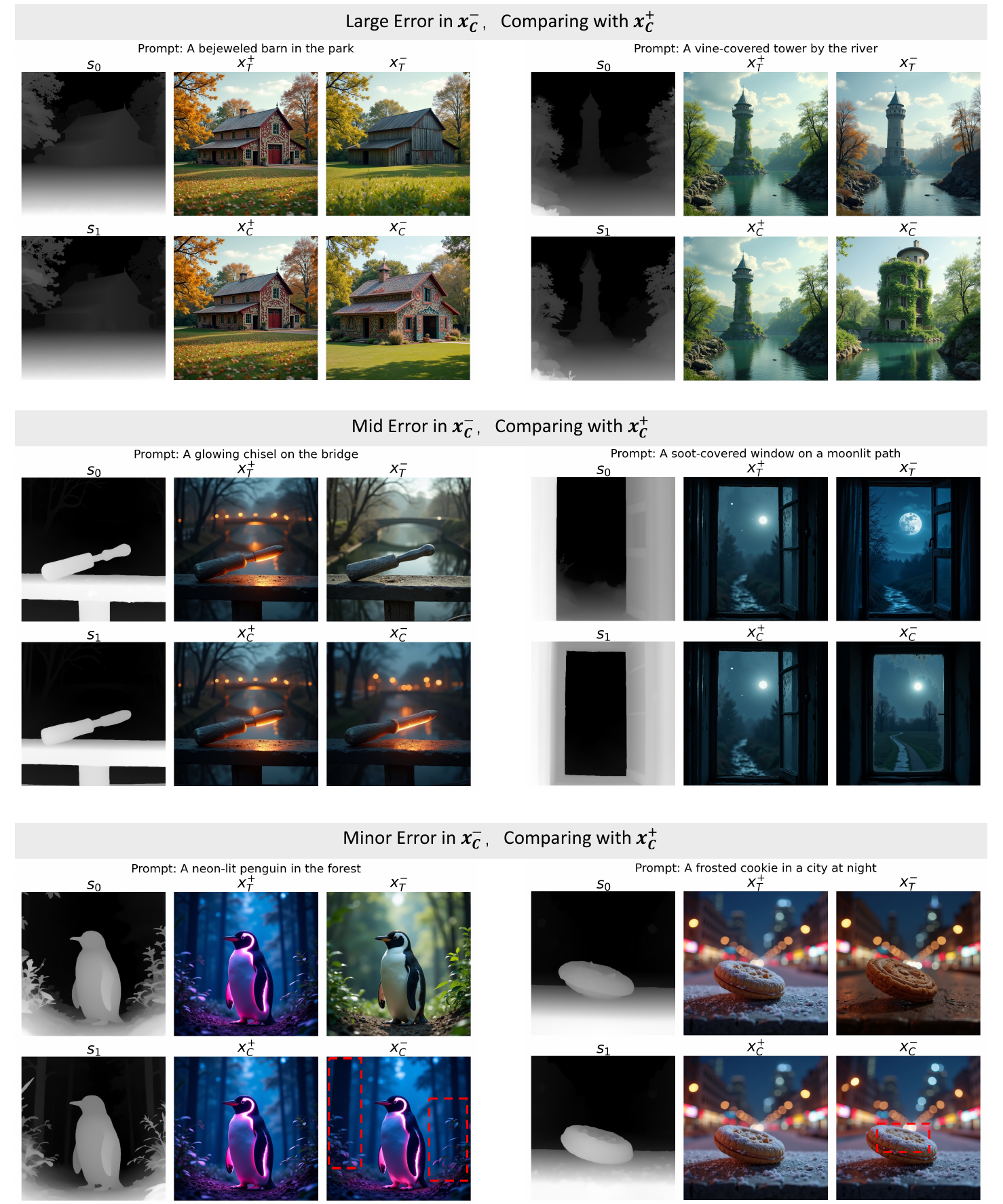}
	\caption{
		\textbf{Additional Examples of Disentangled and Conflict-Aware DPO Preference Data.} 
	}
	\label{fig:more_examples_data}
\end{figure*}

\subsection{Additional Examples of Disentangled and Conflict-Aware DPO Preference Data}

Fig.~\ref{fig:more_examples_data} presents additional examples of our Disentangled and Conflict-Aware DPO data, using depth maps to illustrate our methodology. We construct preference pairs spanning a spectrum of conditional alignment errors—large, mid-level, and minor—to train the model progressively on structural fidelity.

\paragraph{Large Errors} The top row of Fig.~\ref{fig:more_examples_data} shows pairs where the negative sample ($x_C^-$) has significant structural deviations. For instance, the generated ``bejeweled barn'' and ``vine-covered tower'' fail to match the fundamental layout of their depth maps. These examples train the model to capture the global composition.

\paragraph{Mid-level Errors} The middle row presents moderate inconsistencies. The negative samples capture the main objects but err in key aspects, such as the misplaced window in the ``soot-covered window'' scene or the incorrect shape of the ``glowing chisel.'' These pairs refine the model's grasp of spatial relationships.

\paragraph{Minor Errors} The bottom row focuses on fine-grained details. The negative samples are largely faithful but contain subtle inaccuracies, such as ignoring background foliage structures in the ``neon-lit penguin'' scene or failing to render surface grains on the ``frosted cookie.'' These examples hone the model's ability to render precise details, enhancing overall fidelity.

\subsection{Additional Visualizations of Depth, Canny, and Soft-edge Conditions}
\begin{figure*}[ht!]
	\includegraphics[width=1.0\linewidth]{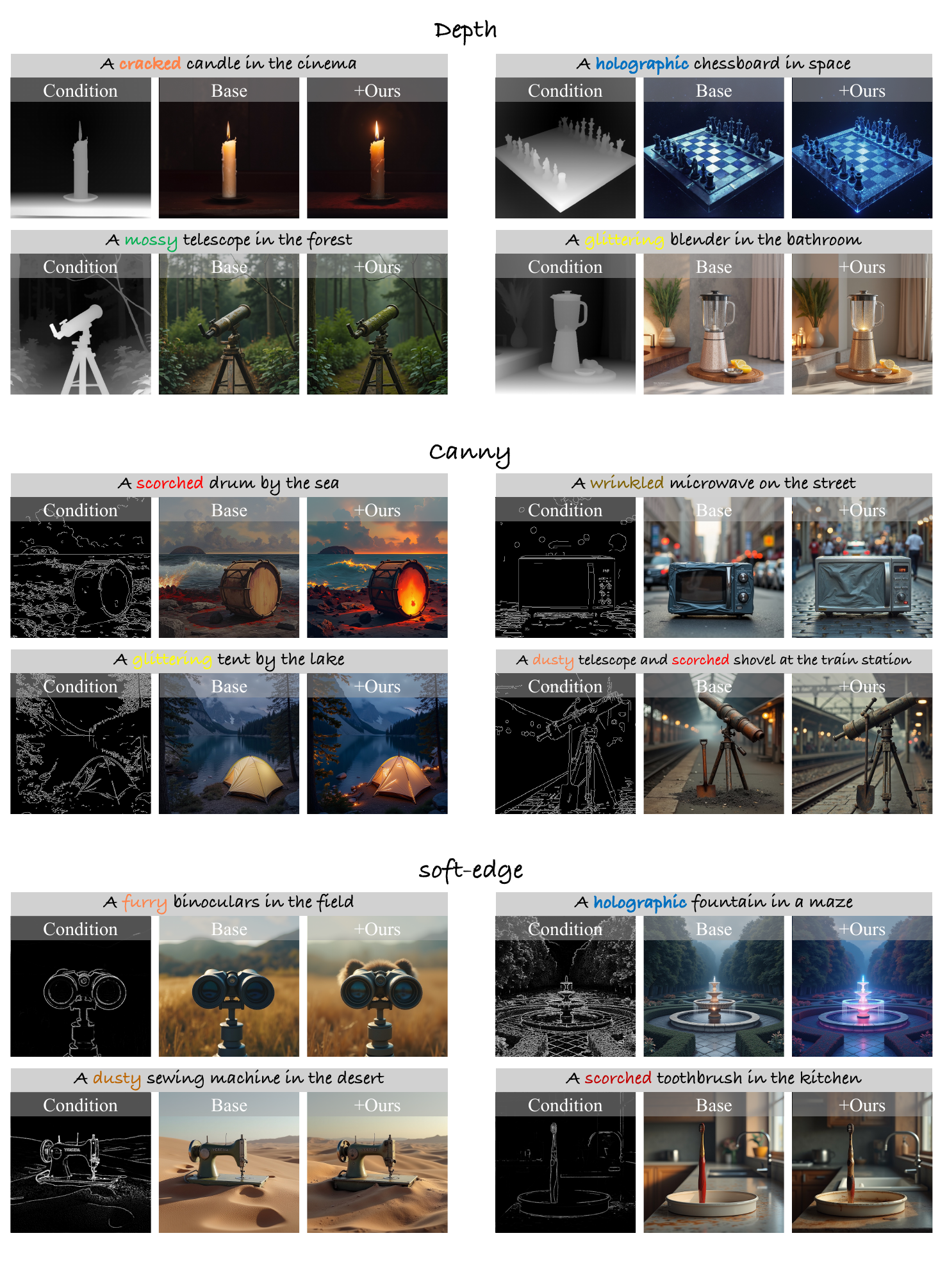}
	\caption{
		\textbf{Additional Examples of Enhancing State-of-the-Art Conditional Image Generation Methods Using Our Approach.} 
	}
	\label{fig:more_vis}
\end{figure*}

In this section, we present comprehensive qualitative results that further demonstrate the superior capabilities of our proposed BideDPO method across diverse conditional image generation scenarios. Fig.~\ref{fig:more_vis} showcases an extensive collection of examples, systematically organized by three distinct conditional modalities, illustrating how our model achieves enhanced fidelity and semantic alignment while maintaining structural integrity across various input conditions.

\subsubsection{Depth Condition}
Fig.~\ref{fig:more_vis} (Top) demonstrates our method's effectiveness when conditioned on depth maps, where grayscale intensity encodes spatial distance information. The BideDPO model exhibits remarkable proficiency in capturing intricate spatial structures while significantly improving semantic alignment with textual prompts compared to baseline methods. Notable improvements include enhanced textural details, such as the subtle crack patterns on candle surfaces and mossy textures on telescope bodies, demonstrating our model's ability to faithfully interpret descriptive adjectives while preserving geometric fidelity.

\subsubsection{Canny Condition}
Fig.~\ref{fig:more_vis} (Middle) showcases results from Canny edge conditions, which demand strict structural adherence. Our method excels at resolving conflicts between textual descriptions and these constraints, rectifying common failures of the baseline model. For instance, it successfully renders challenging attributes such as ``scorched'' or ``glittering''—which the baseline struggles with—thereby significantly enhancing the control and expressive power of Canny-conditioned generation.

\subsubsection{Soft-edge Condition}
Fig.~\ref{fig:more_vis} (Bottom) illustrates our approach under soft-edge conditions, which offer less stringent structural guidance while preserving overall scene composition. BideDPO maintains superior visual quality and semantic relevance even under these more ambiguous constraints, demonstrating the robustness and flexibility of our alignment framework. The results reveal enhanced textural fidelity, such as furry surfaces on binoculars and holographic effects on fountains, showcasing our model's ability to interpret complex descriptive attributes while working within the constraints of softer structural guidance.

\section{Detailed Derivation of Bidirectionally Decoupled DPO}
\label{sec:BideDPO_detail}

In this section, we provide a full step-by-step derivation of our Bidirectionally Decoupled DPO (BideDPO) method, as presented in the main paper.

\subsection{Foundation: Independent Loss Components}
Our method begins by addressing the limitation of vanilla DPO, which uses a single preference pair evaluated under a shared condition. Instead, we define two distinct preference triplets for the text and condition objectives, respectively.

For text alignment, we use the triplet $(x_T^+, x_T^-, c_0)$, where $x_T^+$ is preferred to $x_T^-$ under an initial condition $c_0$. The corresponding text loss is:
\begin{equation}
	\mathcal{L}_{\text{text}} = -\log \sigma\left(\underbrace{f_{\text{text}}(x^+_T, c_0; \theta) - f_{\text{text}}(x^-_T, c_0; \theta)}_{\Delta_{\text{text}}(c_0; \theta)}\right)
	\label{supp:eq_loss_text}
\end{equation}
For condition alignment, we use the triplet $(x_C^+, x_C^-, c_1)$, where $x_C^+$ is preferred to $x_C^-$ under a different, stricter condition $c_1$. The corresponding condition loss is:
\begin{equation}
	\mathcal{L}_{\text{cond}} = -\log \sigma\left(\underbrace{f_{\text{cond}}(x^+_C, c_1; \theta) - f_{\text{cond}}(x^-_C, c_1; \theta)}_{\Delta_{\text{cond}}(c_1; \theta)}\right)
	\label{supp:eq_loss_cond}
\end{equation}

\subsection{Gradient Derivation}
We now derive the partial derivative of the total loss $\mathcal{L}_{\text{decoupled}}$ with respect to the model parameters $\theta$. As noted in the main paper, the adaptive weights are computed with a stop-gradient operator, meaning they are treated as detached constants during the backward pass. Our starting point is the total loss function:
\begin{equation}
	\mathcal{L}_{\text{decoupled}} = w_{\text{text}}\mathcal{L}_{\text{text}} + w_{\text{cond}}\mathcal{L}_{\text{cond}}.
	\label{supp:eq_loss_decoupled}
\end{equation}
Because $w_{\text{text}}$ and $w_{\text{cond}}$ are treated as constants with respect to $\theta$ for the gradient calculation, we can apply the sum rule directly:
\begin{equation}
	\frac{\partial \mathcal{L}_{\text{decoupled}}}{\partial \theta} = w_{\text{text}}\frac{\partial \mathcal{L}_{\text{text}}}{\partial \theta} + w_{\text{cond}}\frac{\partial \mathcal{L}_{\text{cond}}}{\partial \theta}.
	\label{supp:eq_grad_sum}
\end{equation}
Next, we derive the gradients for the individual loss components using the chain rule. For a general loss of the form $L = -\log\sigma(z)$, its derivative is $\frac{\partial L}{\partial \theta} = -(1-\sigma(z))\frac{\partial z}{\partial \theta}$.

Applying this to the text loss component from Eq. \ref{supp:eq_loss_text}:
\begin{equation}
	\frac{\partial \mathcal{L}_{\text{text}}}{\partial \theta} = -\left(1 - \sigma(\Delta_{\text{text}}(c_0; \theta))\right) \frac{\partial \Delta_{\text{text}}(c_0; \theta)}{\partial \theta}
	\label{supp:eq_grad_text}
\end{equation}
And similarly for the condition loss component from Eq. \ref{supp:eq_loss_cond}:
\begin{equation}
	\frac{\partial \mathcal{L}_{\text{cond}}}{\partial \theta} = -\left(1 - \sigma(\Delta_{\text{cond}}(c_1; \theta))\right) \frac{\partial \Delta_{\text{cond}}(c_1; \theta)}{\partial \theta}
	\label{supp:eq_grad_cond}
\end{equation}
Finally, substituting Eqs. \ref{supp:eq_grad_text} and \ref{supp:eq_grad_cond} back into Eq. \ref{supp:eq_grad_sum} gives us the final gradient for our decoupled objective, as presented in the main paper:
\begin{equation}
	\frac{\partial \mathcal{L}_{\text{decoupled}}}{\partial \theta} =
	-w_{\text{text}} \left(1 - \sigma\bigl(\Delta_{\text{text}}(c_0; \theta)\bigr)\right)
	\frac{\partial \Delta_{\text{text}}(c_0; \theta)}{\partial \theta}
	-w_{\text{cond}} \left(1 - \sigma\bigl(\Delta_{\text{cond}}(c_1; \theta)\bigr)\right)
	\frac{\partial \Delta_{\text{cond}}(c_1; \theta)}{\partial \theta}
	\label{supp:eq_grad_final}
\end{equation}

\subsection{Final BideDPO Objective for Diffusion Models}
The general BideDPO framework can be specifically instantiated for diffusion models by defining the preference as a reward based on denoising performance. Our final objective combines the principles of decoupled losses and adaptive balancing, starting from the loss function presented in the main paper:
\begin{equation}
	\mathcal{L}_{\text{BideDPO}}(\theta) = - \mathbb{E}_{\substack{(x_T^+, x_T^-, c_0) \sim \mathcal{D}_T, (x_C^+, x_C^-, c_1) \sim \mathcal{D}_C}} \Big[
		w_{\text{text}} \log\sigma(\beta T \Delta R_T)
		+ w_{\text{cond}} \log\sigma(\beta T \Delta R_C)
		\Big].
\end{equation}

Here, $\Delta R_T$ and $\Delta R_C$ are the total reward differences for the text and condition preference pairs. And $(x_T^+, x_T^-, c_0) \sim \mathcal{D}_T, (x_C^+, x_C^-, c_1) \sim \mathcal{D}_C$ means that the samples are drawn from the text-disentangled and condition-disentangled preference pairs of the same unified preference set $\mathcal{D} = \text{zip}(\mathcal{D}_T, \mathcal{D}_C)$. We first define a per-sample reward $r(x_t, c, \epsilon; \theta)$ as the reduction in denoising error achieved by our model $\epsilon_\theta$ compared to a reference model $\epsilon_{\text{ref}}$:
\begin{equation}
	r(x_t,c,\epsilon;\theta)
	= \lVert\epsilon - \epsilon_{\mathrm{ref}}(x_t,c)\rVert^2
	- \lVert\epsilon - \epsilon_\theta(x_t,c)\rVert^2.
\end{equation}
The total reward differences are then calculated by comparing the rewards of the preferred and dispreferred samples for both the text-aligned pair (under condition $c_0$) and the condition-aligned pair (under condition $c_1$):
\begin{align}
	R_T & = r(x_{t,T}^+,c_0,\epsilon_T^+;\theta) - r(x_{t,T}^-,c_0,\epsilon_T^-;\theta), \\
	R_C & = r(x_{t,C}^+,c_1,\epsilon_C^+;\theta) - r(x_{t,C}^-,c_1,\epsilon_C^-;\theta).
\end{align}

By substituting the definition of the reward function $r(\cdot)$ into these expressions, we can expand them to show the full formulation. For the text-aligned reward difference $R_T$:
\begin{align}
	R_T
	 & = \bigl[\lVert\epsilon_T^+ - \epsilon_{\mathrm{ref}}(x_{t,T}^+,c_0)\rVert^2
		- \lVert\epsilon_T^+ - \epsilon_\theta(x_{t,T}^+,c_0)\rVert^2\bigr]\notag
	- \bigl[\lVert\epsilon_T^- - \epsilon_{\mathrm{ref}}(x_{t,T}^-,c_0)\rVert^2
	- \lVert\epsilon_T^- - \epsilon_\theta(x_{t,T}^-,c_0)\rVert^2\bigr]\notag      \\
	 & = \lVert\epsilon_T^+ - \epsilon_{\mathrm{ref}}(x_{t,T}^+,c_0)\rVert^2
	- \lVert\epsilon_T^+ - \epsilon_\theta(x_{t,T}^+,c_0)\rVert^2
	- \lVert\epsilon_T^- - \epsilon_{\mathrm{ref}}(x_{t,T}^-,c_0)\rVert^2
	+ \lVert\epsilon_T^- - \epsilon_\theta(x_{t,T}^-,c_0)\rVert^2.
\end{align}
Similarly, for the condition-aligned reward difference $R_C$:
\begin{align}
	R_C
	 & = \bigl[\lVert\epsilon_C^+ - \epsilon_{\mathrm{ref}}(x_{t,C}^+,c_1)\rVert^2
		- \lVert\epsilon_C^+ - \epsilon_\theta(x_{t,C}^+,c_1)\rVert^2\bigr]\notag
	- \bigl[\lVert\epsilon_C^- - \epsilon_{\mathrm{ref}}(x_{t,C}^-,c_1)\rVert^2
	- \lVert\epsilon_C^- - \epsilon_\theta(x_{t,C}^-,c_1)\rVert^2\bigr]\notag      \\
	 & = \lVert\epsilon_C^+ - \epsilon_{\mathrm{ref}}(x_{t,C}^+,c_1)\rVert^2
	- \lVert\epsilon_C^+ - \epsilon_\theta(x_{t,C}^+,c_1)\rVert^2
	- \lVert\epsilon_C^- - \epsilon_{\mathrm{ref}}(x_{t,C}^-,c_1)\rVert^2
	+ \lVert\epsilon_C^- - \epsilon_\theta(x_{t,C}^-,c_1)\rVert^2.
\end{align}
Finally, substituting these fully expanded reward differences back into our main objective yields the complete BideDPO loss function used for training:
\begin{equation}
	\begin{split}
		\mathcal{L}_{\mathrm{BideDPO}}(\theta)
		= -\mathbb{E}\Bigl[ &
			w_{\mathrm{text}}
			\log\sigma\Bigl(\beta T\bigl[
				\lVert\epsilon_T^+ - \epsilon_{\mathrm{ref}}(x_{t,T}^+,c_0)\rVert^2
		- \lVert\epsilon_T^+ - \epsilon_\theta(x_{t,T}^+,c_0)\rVert^2 \\
		                    & \quad
				- \lVert\epsilon_T^- - \epsilon_{\mathrm{ref}}(x_{t,T}^-,c_0)\rVert^2
				+ \lVert\epsilon_T^- - \epsilon_\theta(x_{t,T}^-,c_0)\rVert^2
		\bigr]\Bigr)                                                  \\
		                    & + w_{\mathrm{cond}}
			\log\sigma\Bigl(\beta T\bigl[
				\lVert\epsilon_C^+ - \epsilon_{\mathrm{ref}}(x_{t,C}^+,c_1)\rVert^2
		- \lVert\epsilon_C^+ - \epsilon_\theta(x_{t,C}^+,c_1)\rVert^2 \\
		                    & \quad
				- \lVert\epsilon_C^- - \epsilon_{\mathrm{ref}}(x_{t,C}^-,c_1)\rVert^2
				+ \lVert\epsilon_C^- - \epsilon_\theta(x_{t,C}^-,c_1)\rVert^2
				\bigr]\Bigr)
			\Bigr].
	\end{split}
\end{equation}

\section{Additional Discussions}
\subsection{Discussion on More DPO Variants}
\label{sec:discussion_DPO}

In our main experiments, we compare BideDPO against a ``naive'' application of DPO, as described in the Baseline Configurations section. This Naive DPO baseline uses a single preference pair format: the positive sample aligns well with both text and condition ($T^+, C^+$), while the negative sample aligns poorly with both ($T^-, C^-$). This setup, while straightforward, does not fully capture the complexity of the alignment problem, where conflicts can arise from either text or condition independently.

To provide a more robust comparison, we introduce an additional baseline, ``DPO (Mixed)''. In this setting, the negative samples are constructed from a mix of failure cases: poor text and poor condition ($T^-, C^-$), good text but poor condition ($T^+, C^-$), and poor text but good condition ($T^-, C^+$). This creates a more diverse and challenging training signal for the DPO model, forcing it to learn a more nuanced reward function.

The results, presented in Tab.~\ref{tab:depth_results_sup}, reveal an interesting trade-off. The DPO (Mixed) baseline achieves a higher Success Ratio (0.73 vs. 0.71) and CLIP score (0.2884 vs. 0.2860) compared to the Naive DPO, indicating improved text alignment. However, it performs worse on conditional fidelity, with higher (worse) MSE and SGMSE scores. This suggests that while a mixed-negative strategy helps the model better understand textual nuances, the undifferentiated DPO loss struggles to balance the competing objectives, leading to a degradation in structural adherence.

In contrast, our BideDPO method significantly outperforms both DPO baselines across all metrics. By explicitly decoupling the preference pairs for text and condition alignment, BideDPO provides clear, unambiguous learning signals for each objective. This, combined with our adaptive weighting mechanism, allows the model to simultaneously improve both text-prompt consistency and conditional fidelity, overcoming the trade-offs that limit standard DPO approaches.

\begin{table}[h!]
	\centering
	\begin{tabular}{rcccc}
        \toprule[1pt]
		\textbf{Method}   & \textbf{SR} $\uparrow$ & \textbf{MSE} $\downarrow$ & \textbf{SGMSE} $\downarrow$ & \textbf{CLIP} $\uparrow$ \\
		\hline
		\rowcolor[HTML]{F8FBFF}
		Union-Pro2        & 0.49                              & 176.982
                   & 272.400                     & 0.2748
                           \\
		\rowcolor[HTML]{F8FBFF}
		+ DPO (Naive)  &                0.71                  &  168.284
                         &          219.935                  & 0.2860
                           \\
        \rowcolor[HTML]{F8FBFF}
        + DPO (Mixed) & 0.73 & 186.417 & 229.857 & 0.2884 \\
		\rowcolor[HTML]{F8FBFF}
		+ Ours & \textbf{0.84}                     & \textbf{163.968}           & \textbf{195.728}             & \textbf{0.2924} \\
		\toprule[1pt]
	\end{tabular}
    \vspace{-2mm}
	\caption{Comparison of different DPO configurations for depth-conditioned image generation. Our method surpasses both naive and mixed DPO baselines.}
    \vspace{-4mm}
	\label{tab:depth_results_sup}
\end{table}

\subsection{\rebuttal{Comparison with More Image Conditional Generation Methods}}
\label{sec:more_image_conditional_comparison}

\rebuttal{To evaluate BideDPO against state-of-the-art approaches for conditional image generation, we compare with LooseControl~\citep{bhat2024loosecontrol} and ControlNet++~\citep{li2024controlnet++}.}

\subsubsection{\rebuttal{Results on DualAlign Benchmark}}

\rebuttal{Tab.~\ref{tab:more_image_conditional_dualalign} presents the quantitative comparison on the DualAlign benchmark with depth conditioning. BideDPO achieves significantly superior performance across all metrics, with a success rate (SR) of \textbf{0.84}, MSE of \textbf{164.0}, SGMSE of \textbf{195.7}, and CLIP score of \textbf{0.2924}. In contrast, LooseControl and ControlNet++ show limited performance, with SR values of 0.43 and 0.49, respectively, and significantly higher structural errors (MSE: 791.13 and 331.85).}

\subsubsection{\rebuttal{Results on COCO Benchmark}}

\rebuttal{Tab.~\ref{tab:more_image_conditional_coco} shows the comparison on the COCO benchmark with depth conditioning. BideDPO again achieves superior performance, with an SR of \textbf{0.91}, MSE of \textbf{236.3}, SGMSE of \textbf{245.3}, and CLIP score of \textbf{0.2633}. LooseControl and ControlNet++ show lower success rates (0.72 and 0.79) and significantly higher structural errors.}

\subsubsection{\rebuttal{Discussion}}

\rebuttal{The comparison reveals that existing approaches alone are insufficient to resolve conflicts between text and condition constraints. In contrast, BideDPO's post-training approach with bidirectionally decoupled objectives enables it to effectively harmonize both constraints without sacrificing either objective.}

\begin{table}[t]
\centering
\caption{\rebuttal{\textbf{Comparison with more image conditional generation methods on DualAlign benchmark.}}}
\vspace{-4mm}
\footnotesize
\setlength{\tabcolsep}{1.8mm}
\begin{tabular}{lcccc}
\toprule[1pt]
\textbf{Method} & \textbf{SR} $\uparrow$ & \textbf{MSE} $\downarrow$ & \textbf{SGMSE} $\downarrow$ & \textbf{CLIP} $\uparrow$ \\
\midrule
\rowcolor[HTML]{F5F5F5}
LooseControl & 0.43 & 791.13 & 1280.17 & 0.2852 \\
\rowcolor[HTML]{F5F5F5}
ControlNet++ & 0.49 & 331.85 & 480.66 & 0.2854 \\
\rowcolor[HTML]{FFF8F0}
\textbf{BideDPO (ours)} & \textbf{0.84} & \textbf{164.0} & \textbf{195.7} & \textbf{0.2924} \\
\bottomrule[1pt]
\end{tabular}
\label{tab:more_image_conditional_dualalign}
\end{table}

\begin{table}[t]
\centering
\caption{\rebuttal{\textbf{Comparison with more image conditional generation methods on COCO benchmark.}}}
\vspace{-4mm}
\footnotesize
\setlength{\tabcolsep}{1.8mm}
\begin{tabular}{lcccc}
\toprule[1pt]
\textbf{Method} & \textbf{SR} $\uparrow$ & \textbf{MSE} $\downarrow$ & \textbf{SGMSE} $\downarrow$ & \textbf{CLIP} $\uparrow$ \\
\midrule
\rowcolor[HTML]{F5F5F5}
LooseControl & 0.72 & 1334.01 & 1705.97 & 0.2534 \\
\rowcolor[HTML]{F5F5F5}
ControlNet++ & 0.79 & 548.26 & 668.92 & 0.2557 \\
\rowcolor[HTML]{FFF8F0}
\textbf{BideDPO (ours)} & \textbf{0.91} & \textbf{236.3} & \textbf{245.3} & \textbf{0.2633} \\
\bottomrule[1pt]
\end{tabular}
\label{tab:more_image_conditional_coco}
\end{table}

\subsection{\rebuttal{Comparison with DPO-based Post-Training Methods}}
\label{sec:dpo_comparison}

\rebuttal{To evaluate BideDPO against state-of-the-art DPO-based post-training methods for conditional image generation, we compare with SPO~\citep{liang2024spo} and RankDPO~\citep{rankdpo}. These methods are based on DPO with some improvements for preference optimization, but differ from our approach in how they handle the dual objectives of text alignment and conditional fidelity. Note that RankDPO is not publicly available, so we implement it based on the methodology described in the paper. All methods share the same FLUX backbone and evaluation pipeline to ensure fair comparison.}

\subsubsection{\rebuttal{Results on DualAlign Benchmark}}

\rebuttal{Tab.~\ref{tab:dpo_dualalign} presents the quantitative comparison on the DualAlign benchmark with depth conditioning. BideDPO achieves the best performance across all metrics, with a success rate (SR) of \textbf{0.84}, MSE of \textbf{164.0}, SGMSE of \textbf{195.7}, and CLIP score of \textbf{0.2924}.}

\rebuttal{SPO achieves a competitive SR of 0.78 with good structural control (MSE: 166.2, SGMSE: 208.7), while RankDPO reaches a SR of 0.83 but suffers from higher structural errors (MSE: 188.5, SGMSE: 235.6). Both methods struggle to fully resolve conflicts between text and condition constraints.}

\rebuttal{In contrast, BideDPO's bidirectionally decoupled objective effectively balances both text alignment and structural conditioning by explicitly separating preference pairs along text and condition axes, enabling the model to simultaneously improve both text-prompt consistency and conditional fidelity.}

\subsubsection{\rebuttal{Discussion}}

\rebuttal{The comparison with DPO-based methods reveals different trade-offs: SPO maintains better structural control but achieves lower text alignment, while RankDPO improves text alignment but struggles with structural fidelity. This suggests that existing DPO-based methods, while effective, do not fully address the challenge of simultaneously optimizing both text and condition objectives when they conflict.}

\rebuttal{BideDPO's explicit decoupling of text and condition objectives, combined with adaptive loss balancing, enables it to outperform both baselines by effectively harmonizing both constraints without sacrificing either objective. The bidirectionally decoupled approach provides clearer learning signals for each objective, allowing the model to learn more effectively from preference pairs that may be ambiguous when both objectives are considered together.}

\begin{table}[t]
\centering
\caption{\rebuttal{\textbf{Comparison with DPO-based post-training methods on DualAlign benchmark.} All methods share the same FLUX backbone and evaluation pipeline.}}
\vspace{-4mm}
\footnotesize
\setlength{\tabcolsep}{1.8mm}
\begin{tabular}{lcccc}
\toprule[1pt]
\textbf{Method} & \textbf{SR} $\uparrow$ & \textbf{MSE} $\downarrow$ & \textbf{SGMSE} $\downarrow$ & \textbf{CLIP} $\uparrow$ \\
\midrule
\rowcolor[HTML]{F8FBFF}
SPO & 0.78 & 166.2 & 208.7 & 0.2881 \\
\rowcolor[HTML]{F8FBFF}
RankDPO & 0.83 & 188.5 & 235.6 & 0.2914 \\
\rowcolor[HTML]{FFF8F0}
\textbf{BideDPO (Ours)} & \textbf{0.84} & \textbf{164.0} & \textbf{195.7} & \textbf{0.2924} \\
\bottomrule[1pt]
\end{tabular}
\label{tab:dpo_dualalign}
\end{table}

\subsection{\rebuttal{Stable Diffusion 1.5 + BideDPO}}
\label{sec:sd15_bidedpo}
\begin{figure*}[ht!]
	\includegraphics[width=1.0\linewidth]{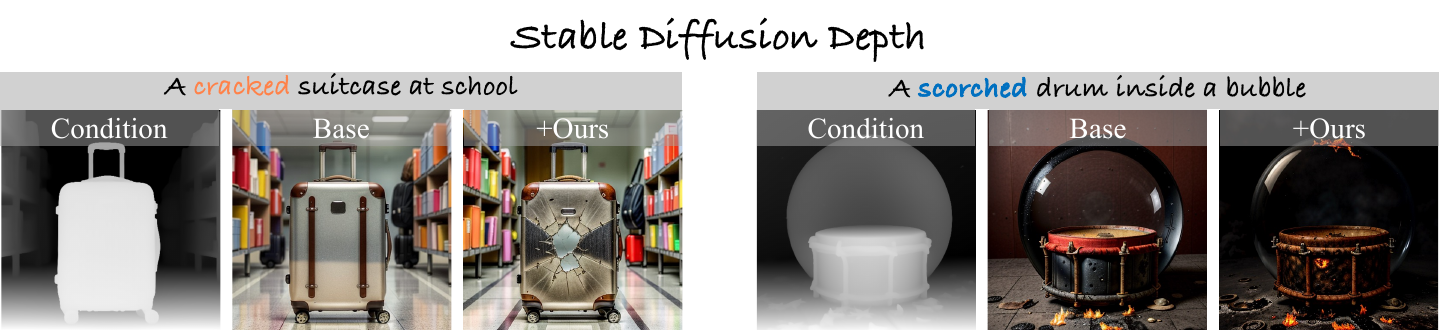}
	\caption{
		\rebuttal{\textbf{Stable Diffusion 1.5 depth-conditioned generation on DualAlign.}
		Qualitative comparison between the ControlNet baseline and our BideDPO fine-tuned model.
		BideDPO preserves the textual semantics while aligning more faithfully with the provided depth controls, yielding sharper geometry and cleaner spatial layouts.}
	}
	\label{fig:sd15_depth_vis}
\end{figure*}

\noindent\textbf{\rebuttal{Stable Diffusion 1.5 + BideDPO.}}
\rebuttal{To demonstrate that BideDPO is not limited to FLUX-based models but also generalizes to Stable Diffusion-based architectures, we further fine-tune the ControlNet of Stable Diffusion 1.5 with our bidirectionally decoupled objective. To highlight how BideDPO improves controllable generation under the DualAlign depth benchmark, we compare our approach with the ControlNet baseline. Relative to ControlNet, our model dramatically increases success rate while simultaneously lowering both MSE and SGMSE, indicating superior adherence to depth conditioning without sacrificing semantic fidelity. Notably, the CLIP score also rises, underscoring that the additional control signal does not compromise text alignment. The detailed quantitative comparison is provided in Tab.~\ref{tab:depth_sd15_results}, while Fig.~\ref{fig:sd15_depth_vis} visualizes the accompanying qualitative gains.}

\begin{table}[t]
  \centering
\caption{\rebuttal{\textbf{Stable Diffusion 1.5 depth-conditioned image generation on DualAlign Benchmark.} ``Ctrl.'' indicates support for conditional generation; qualitative comparisons appear in Fig.~\ref{fig:sd15_depth_vis}.}}
  \label{tab:depth_sd15_results}
  \footnotesize
  \setlength{\tabcolsep}{0.8mm}
  \begin{tabular}{rccccc}
    \toprule[1pt]
    \textbf{Method} & \textbf{Ctrl.} & \textbf{SR} $\uparrow$ & \textbf{MSE} $\downarrow$ & \textbf{SGMSE} $\downarrow$ & \textbf{CLIP} $\uparrow$ \\
    \hline
    \rowcolor[HTML]{F5F5F5}
    SD 1.5-ControlNet & $\checkmark$ & 0.50 & 391.80 & 592.92 & 0.2824 \\
    \rowcolor[HTML]{F8FBFF}
    + BideDPO (Ours) & $\checkmark$ & \textbf{0.71} & \textbf{187.44} & \textbf{234.14} & \textbf{0.2853} \\
    \toprule[1pt]
  \end{tabular}
\end{table}

\subsection{\rebuttal{Multi-Condition Generation: Text + Depth + Canny}}
\label{sec:multi_cond}
\begin{figure*}[ht!]
	\includegraphics[width=1.0\linewidth]{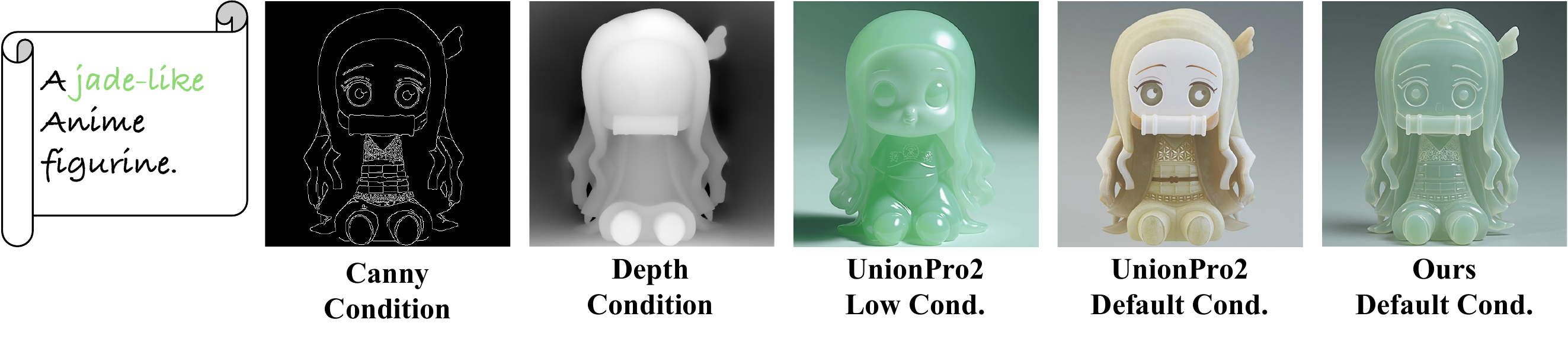}
	\caption{
		\rebuttal{\textbf{Multi-condition generation with text, depth, and Canny edge controls.}
		Example demonstrating BideDPO's capability to handle multiple simultaneous conditioning inputs. The text prompt is ``A jade-like Anime figurine.'' We condition the generation on both depth maps and Canny edges extracted from the original image, while applying the jade-like style transformation. Our method (rightmost) successfully harmonizes all three constraints—textual semantics, depth geometry, and edge structure—producing a high-fidelity jade-like figurine that preserves both spatial layout and fine-grained details. In contrast, UnionPro2 struggles to balance these competing objectives, either losing structural fidelity (low cond) or failing to fully realize the style transformation (default cond).}
	}
	\label{fig:multi_cond_vis}
\end{figure*}

\noindent\textbf{\rebuttal{Scaling to Multiple Simultaneous Conditions.}}
\rebuttal{To demonstrate that BideDPO scales beyond two conditioning modalities, we showcase a challenging multi-condition generation scenario that simultaneously enforces text prompts, depth maps, and Canny edge constraints. As illustrated in Fig.~\ref{fig:multi_cond_vis}, we condition the generation on both depth and Canny edge maps extracted from an original anime figurine image, while applying the text prompt ``A jade-like Anime figurine.'' This setup creates a complex multi-objective optimization problem where the model must harmonize three distinct constraints: (1) textual semantics (jade-like material transformation), (2) depth geometry (preserving 3D spatial structure), and (3) edge structure (maintaining fine-grained boundaries and details).}

\rebuttal{Our BideDPO method successfully balances all three objectives, producing a high-quality jade-like figurine that preserves both the global spatial layout from the depth map and the fine-grained details captured by the Canny edges, while faithfully realizing the translucent, polished jade aesthetic described in the text prompt. In contrast, UnionPro2 struggles to simultaneously satisfy all constraints: with low conditioning strength, it loses structural fidelity to the depth and edge maps; with default conditioning strength, it maintains better structural alignment but fails to fully realize the style transformation, resulting in a less convincing jade-like appearance. This example demonstrates that BideDPO's decoupled objective and adaptive loss balancing mechanism naturally extend to handle multiple conditioning inputs by treating each conditioning path independently and dynamically adjusting their relative importance during training.}

\begin{table}[t]
  \centering
  \caption{\rebuttal{\textbf{Results for multi-condition generation (depth + canny) on DualAlign Benchmark.} The model is simultaneously conditioned on both depth and canny edge maps. Left half shows depth-conditioned results, right half shows canny-conditioned results.}}
  \label{tab:multi_cond_results}
  \vspace{-4mm}
  \footnotesize
  \setlength{\tabcolsep}{0.8mm}
  \begin{tabular}{r|cccc|cccc}
    \toprule[1pt]
    & \multicolumn{4}{c|}{\textbf{Depth Benchmark}} & \multicolumn{4}{c}{\textbf{Canny Benchmark}} \\
    \cmidrule(lr){1-5} \cmidrule(lr){6-9}
    \textbf{Method} & \textbf{SR} $\uparrow$ & \textbf{MSE} $\downarrow$ & \textbf{SGMSE} $\downarrow$ & \textbf{CLIP} $\uparrow$ & \textbf{SR} $\uparrow$ & \textbf{F1} $\uparrow$ & \textbf{SG F1} $\uparrow$ & \textbf{CLIP} $\uparrow$ \\
    \hline
    \rowcolor[HTML]{F8FBFF}
    Union-Pro2 & 0.49 & 177.0 & 272.4 & 0.2748 & 0.34 & 0.418 & 0.143 & 0.2753 \\
    \rowcolor[HTML]{F8FBFF}
    + Ours (depth+canny merge) & \textbf{0.81} & \textbf{159.6} & \textbf{197.5} & \textbf{0.2901} & \textbf{0.68} & \textbf{0.594} & \textbf{0.381} & \textbf{0.2857} \\
    \toprule[1pt]
  \end{tabular}
\end{table}

\rebuttal{Tab.~\ref{tab:multi_cond_results} provides quantitative evidence that BideDPO effectively handles simultaneous depth and Canny edge conditioning. When evaluated on the DualAlign depth benchmark, our method achieves a success rate (SR) of 0.81, significantly outperforming Union-Pro2's 0.49, while simultaneously reducing both MSE (159.6 vs. 177.0) and SGMSE (197.5 vs. 272.4) errors, indicating superior structural fidelity. The CLIP score also improves from 0.2748 to 0.2901, demonstrating enhanced text alignment. On the Canny benchmark, BideDPO maintains strong performance with SR of 0.68 (vs. 0.34 for Union-Pro2), F1 score of 0.594 (vs. 0.418), and SG F1 of 0.381 (vs. 0.143), while improving CLIP from 0.2753 to 0.2857. These results confirm that BideDPO's bidirectional decoupling mechanism successfully harmonizes multiple conditioning modalities without compromising performance on either benchmark, validating the method's scalability to complex multi-condition generation scenarios.}

\subsection{\rebuttal{Discussion on Adaptive Loss Balancing (ALB) Methods}}
\label{sec:discussion_alb}

\rebuttal{We investigate the robustness of ALB to different weight calculation methods and batch sizes.}

\subsubsection{\rebuttal{Instance Mean vs. Historical Mean}}

\rebuttal{We compare our default instance mean approach with a historical mean (moving average) approach. As shown in Tab.~\ref{tab:alb_comparison}, both methods achieve similar performance (SR: \textbf{0.84}), with minimal differences across metrics. This robustness stems from our normalization scheme that prevents unstable weight fluctuations, making the simpler instance mean approach a practical choice.}

\subsubsection{\rebuttal{Batch Size Sensitivity}}

\rebuttal{We evaluate ALB with batch sizes of 8, 16, and 32. Tab.~\ref{tab:alb_comparison} shows consistent performance across all batch sizes (SR: 0.83-0.85), demonstrating that ALB maintains effective loss balancing regardless of batch size.}

\begin{table}[h]
\centering
\caption{\rebuttal{Comparison of different ALB methods and batch size sensitivity.}}
\label{tab:alb_comparison}
\begin{tabular}{lcccc}
\toprule
Method & SR $\uparrow$ & MSE $\downarrow$ & SGMSE $\downarrow$ & CLIP $\uparrow$ \\
\midrule
ALB instance mean (batch=8) & \textbf{0.84} & \textbf{164.0} & \textbf{195.7} & 0.2924 \\
ALB instance mean (batch=16) & 0.83 & 167.4 & 199.2 & \textbf{0.2940} \\
ALB instance mean (batch=32) & \textbf{0.85} & 160.7 & 195.5 & 0.2900 \\
\midrule
ALB historical mean & 0.84 & 166.2 & 198.8 & \textbf{0.2934} \\
\bottomrule
\end{tabular}
\end{table}

\subsection{Discussion on the Number of Iterations}
\label{sec:discussion_iterations}

As shown in the ablation study in the main paper (Tab.~\ref{tab:iter_results}), the performance of our method improves steadily from the baseline up to Iteration 3, which we selected as our final model. However, we observed a slight degradation in performance at Iteration 4. This phenomenon suggests a potential for overfitting and highlights the trade-offs inherent in our iterative optimization strategy.

We hypothesize that this performance drop is due to the model beginning to overfit to the biases of our automated data generation and scoring pipeline. While the iterative process is highly effective at bootstrapping performance, it creates a feedback loop where the model is trained exclusively on data it generates itself. After several iterations, the data distribution, while high-quality, may become narrower and reflect the specific quirks of the generator and the VLM used for scoring.

At Iteration 4, the model may start to fit to these artifacts rather than learning a more generalizable representation of text and condition alignment. The preference pairs generated may also become less informative, as the distinction between ``preferred'' and ``dispreferred'' samples becomes increasingly subtle for an already powerful generator. Iteration 3 appears to represent the optimal balance point, where the model has reaped the benefits of high-quality, self-generated data without yet succumbing to the effects of overfitting to its own narrowing data distribution.

\subsection{\rebuttal{Detailed Discussion on VLM Selection}}
\label{sec:discussion_vlm}

\rebuttal{In our evaluation pipeline, we employ Qwen2.5-VL-72B as the primary VLM for assessing text-image alignment through the SR metric. This choice warrants careful justification, as the reliability of our conclusions depends on the quality and consistency of the evaluator.}

\rebuttal{\textbf{Rationale for Qwen2.5-VL-72B.} We deliberately adopt Qwen2.5-VL-72B as our primary judge because it is fully open-source and freely usable, and has become a de-facto standard VLM in recent academic work on text--image evaluation. This makes our pipeline easier to reproduce and our SR metric easier to compare against future papers---even if Qwen is not always the single most SOTA model on every benchmark. In addition, Qwen offers strong coverage across diverse object categories and scene types, which is important for our broad DualAlign setting. Its accessibility and community acceptance make it a practical, ``standard'' evaluator that lowers the barrier for future work to build on our method.}

\rebuttal{\textbf{Cross-VLM and Human Validation.} To ensure that our conclusions are not artifacts of a specific VLM, we re-evaluate the same test set with GPT-4o and human raters. The results demonstrate strong consistency across all evaluators:} 
\rebuttal{As shown in Tab.~\ref{tab:vlm_validation_main_paper}, all three evaluators consistently rank $+Ours > +DPO > +SFT > UnionPro2$, with BideDPO achieving the highest scores across all judges (0.82--0.84). Notably, Qwen and Human evaluators both assign \textbf{0.84} to our method, demonstrating that Qwen serves as a reliable proxy for human judgment. The consistent relative rankings across different evaluators validate that our conclusions are robust and not artifacts of a specific VLM, confirming that using Qwen as a representative VLM evaluator is appropriate.}

\end{document}